\documentclass[runningheads,dvipsnames]{llncs}
\usepackage{graphicx}
\usepackage{bbm}
\usepackage{epsfig}
\usepackage{tikz}
\usepackage{comment}
\usepackage{amsmath,amssymb} %
\usepackage{color}
\usepackage{xcolor}
\usepackage{colortbl}
\usepackage{tabulary}
\usepackage{array}
\usepackage{rotating}
\usepackage{booktabs}
\usepackage{breakcites}
\usepackage{capt-of}  %
\usepackage{multirow} %
\usepackage{multicol}
\usepackage{orcidlink} %
\usepackage{pifont}
\usepackage{siunitx}

\usepackage[accsupp]{axessibility}

\usepackage[capitalise]{cleveref}

\newcommand{\PAR}[1]{\noindent{\bf #1~}}

\newcommand{\cmark}{\ding{51}}
\newcommand{\xmark}{\ding{55}}

\newcommand{\R}{\mathbb{R}}
\newcommand{\1}{\mathbbm{1}}
\DeclareMathOperator*{\argmin}{arg\,min}

\def\vec#1{\mathchoice%
	{\mbox{\boldmath $\displaystyle\bf#1$}}
	{\mbox{\boldmath $\textstyle\bf#1$}}
	{\mbox{\boldmath $\scriptstyle\bf#1$}}
	{\mbox{\boldmath $\scriptscriptstyle\bf#1$}}}

\def\mat#1{\mathchoice{\mbox{\boldmath$\displaystyle\tt#1$}}
	{\mbox{\boldmath$\textstyle\tt#1$}}
	{\mbox{\boldmath$\scriptstyle\tt#1$}}
	{\mbox{\boldmath$\scriptscriptstyle\tt#1$}}}

\newcommand{\newtext}[1]{{#1\xspace}} %
 
\newcommand{\camr}[1]{{#1\xspace}}
\newcommand{\hl}[1]{{#1\xspace}}

\definecolor{cgreen}{RGB}{237, 252, 233}
\definecolor{cgrey}{RGB}{241, 241, 243}

\definecolor{cdgreen}{RGB}{30, 128, 0}
\definecolor{cdgrey}{RGB}{51, 51, 51}
\definecolor{csky}{RGB}{51, 204, 255}
\definecolor{cpink}{RGB}{255, 77, 166}
\definecolor{cyel}{RGB}{255, 238, 186}%

\begin{document}
\pagestyle{headings}
\mainmatter
\def\ECCVSubNumber{6212}  %

\title{Is Geometry Enough for Matching in Visual Localization?
} %

\titlerunning{Is Geometry Enough for Matching in Visual Localization?}
\author{Qunjie Zhou\thanks{Equal contribution.}\inst{1}\orcidlink{0000-0002-2434-2393} \and
Sérgio Agostinho$^\star$\inst{2}\orcidlink{0000-0001-7008-1756} \and
Aljoša Ošep\inst{1}\orcidlink{0000-0001-8105-4737} \and
Laura Leal-Taixé\inst{1}\orcidlink{0000-0001-8709-1133}
}
\authorrunning{Q. Zhou et al.}

\institute{$^1$Technical University of Munich, Germany 
\quad
$^2$Universidade de Lisboa, Portugal
\\
\email{$^1$\{qunjie.zhou,aljosa.osep,leal.taixe\}@tum.de}
\quad
\email{$^2$sergio.agostinho@tecnico.ulisboa.pt}
\\
\small{\texttt{
\href{https://github.com/dvl-tum/gomatch}{https://github.com/dvl-tum/gomatch}
}}}

\maketitle

\begin{center}
 \includegraphics[width=1.0\textwidth]{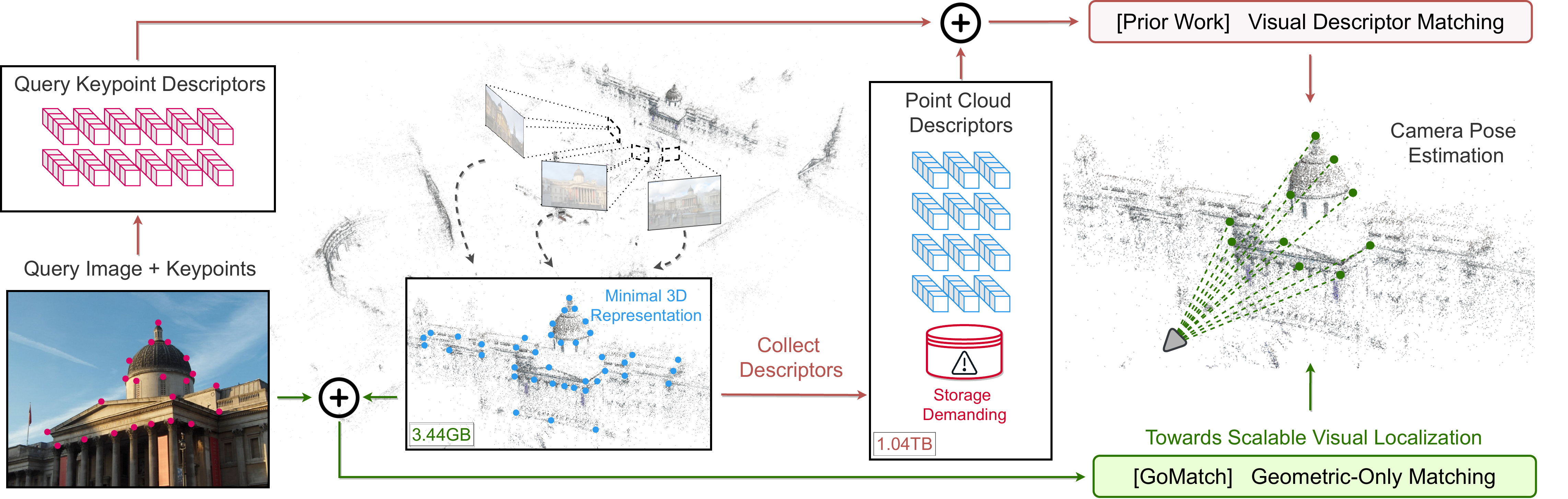}
 \captionof{figure}{In this work, we propose GoMatch to tackle visual localization \wrt a scene represented as a 3D point cloud. By relying only on geometric information for matching, GoMatch allows structure-based methods to achieve localization solely through the use of keypoints, sidestepping the need to store visual descriptors for matching. \newtext{Keeping only the minimal representation of a 3D model, \ie, its coordinates, leads to a more scalable pipeline towards large-scale localization that bypasses privacy concerns and is easy to maintain.} 
}
 \label{fig:teaser}
\end{center}%

\begin{abstract}

\newtext{
In this paper, we propose to go beyond the well-established approach to vision-based localization that relies on visual descriptor matching between a query image and a 3D point cloud. 
While matching keypoints via visual descriptors makes localization highly accurate, it has significant storage demands, raises privacy concerns and
\camr{requires update to the descriptors in the long-term}.
To elegantly address those practical challenges for large-scale localization, we present GoMatch, an alternative to \textit{visual-based matching} that solely relies on geometric information for matching image keypoints to maps, represented as sets of bearing vectors.
Our novel bearing vectors representation of 3D points, significantly relieves the cross-modal challenge in \textit{geometric-based matching} that prevented prior work to tackle localization in a realistic environment.
With additional careful architecture design, GoMatch improves over prior geometric-based matching work with a reduction of ($10.67m, 95.7^{\circ}$) and ($1.43m$, $34.7^{\circ}$) in average median pose errors on Cambridge Landmarks and 7-Scenes, while requiring as little as $1.5/1.7\%$ of storage capacity in comparison to the best visual-based matching methods. 
This confirms its potential and feasibility for real-world localization and opens the door to future efforts in advancing city-scale visual localization methods that do not require storing visual descriptors.
}

\end{abstract}

\section{Introduction}\label{sec:intro}
In this paper we tackle scalable, data-driven visual localization. The ability to localize a query image within a 3D map based representation of the environment is vital in many applications, ranging from robotics to virtual and augmented reality.
In past years, researchers have made a significant progress in vision-based localisation~\cite{Dusmanu2019CVPR,Detone2018CVPRW,Wang2020ECCV,Luo2020CVPR,Muja2014PAMI,Sarlin2020CVPR,Germain2020ECCV,Zhang2019ICCV,Sun2020CVPR,Sattler2018CVPR}. The majority of methods~\cite{Toft2020TPAMI,Dusmanu2019CVPR,Wang2020ECCV,Sarlin2020CVPR,Sun2020CVPR} rely on a pre-built 3D representation of the environment, typically obtained using structure-from-motion (SfM) techniques~\cite{Schoenberger2016CVPR,Schonberger2016ECCV}. Such 3D maps store 3D points and \mbox{$D$-dimensional} visual feature descriptors~\cite{Sattler2017CVPR}. 
To determine the pose of a query image, \ie, its 3D position and orientation, these methods match visual descriptors, obtained from the query image, with the ones stored in the point cloud.
Once image-to-point-cloud matches are established, a Perspective-n-Point (PnP) solver~\cite{Kneip2011CVPR,Gao2003TPAMI} is used to estimate the camera pose.
While working well in practice, this approach suffers from several drawbacks. 
First, we need to explicitly store per-point visual descriptors for point clouds, which hinders its applicability to large-scale environments due to the expensive storage requirement. 
\hl{Second, this limits the applicability to point clouds with specific descriptors, which increases the 3D map descriptor maintenance effort -- maps need to be re-built or updated to be used in conjunction with newly developed descriptors~\cite{Dusmanu2021ICCV}}. 
Third, this approach in practice necessitates a visual descriptor exchange between the server (storing the 3D model and descriptors) and an online feature extractor. 
This is a point of privacy vulnerability, as human identities and personal information can be recovered from visual descriptors intercepted during the transmission~\cite{Pittaluga2019CVPR,Dosovitskiy2016CVPR,Dosovitskiy2016NeurIPS,Chelani2021CVPR,Dusmanu2020CVPR,Geppert2021CVPR,Speciale2019CVPR,Geppert2020ECCV}.
The aforementioned issues lead to the main question we pose in this paper:  \textit{can we localize an image without relying on visual descriptors?}
This would significantly reduce the map storage demands and \hl{get rid of descriptor maintenance}.
Recently, Campbell \etal~\cite{Campbell2020ECCV,Liu2020arXiv} showed that it is feasible to directly match 2D image keypoints with a 3D point cloud using only geometrical cues. However, this is limited to ideal scenarios where outliers are not present.
This assumption does not hold in real-world scenes and is not directly applicable to challenging visual localization. This is not surprising, as relying only on geometrical cues is a significantly more challenging compared to matching visual descriptors. 
In contrast to a single 2D/3D point coordinate, a visual descriptor provides a rich visual context, since it is commonly extracted from the local image patch centered around a keypoint~\cite{Dusmanu2019CVPR,Detone2018CVPRW,Wang2020ECCV,Luo2020CVPR}. 

In this paper, we achieve significant progress in making keypoints-to-point cloud direct matching ready for real-world visual localization. 
To cope with noisy images, point clouds, and inevitably keypoint outliers, we present \textbf{GoMatch}, a novel neural network architecture that relies on \textbf{G}eometrical information \textbf{o}nly. 
GoMatch leverages self- and cross- attention mechanisms to establish initial correspondences between image keypoints and point clouds, and further improves the matching robustness by filtering match outliers using a classifier.
To the best of our knowledge, GoMatch is the first approach that is applicable to visual localization \camr{\textit{in the wild}} and does not rely on storage-demanding visual descriptors. In particular, compared to its prior work on geometric matching-based localization, \newtext{GoMatch leads to a reduction of ($10.67m, 95.7^{\circ}$) and ($1.43m, 34.7^{\circ}$) in average median pose errors on Cambridge Landmarks dataset~\cite{Kendall2015ICCV} and 7-Scenes dataset~\cite{Shotton2013CVPR}}, confirming its potential in real-world visual localization.

We summarize our contributions as the following: (i) we develop a novel method to match query keypoints to a point cloud relying only on geometrical information; 
(ii) We bridge the difference in data modalities between a 2D image keypoint to a 3D point by representing it with its bearing vectors projected into co-visible reference views and show this is remarkably more robust compared to direct cross-modal matching;
(iii) Our extensive evaluation shows that our method significantly outperforms prior work, effectively enabling real-world visual localization based on geometric-only matching;
(iv) Finally, we thoroughly compare our method to the well-established
visual localization \newtext{baselines} and discuss advantages and disadvantages of each approach.
With this analysis, we hope to open the door for future progress \newtext{towards} more general and scalable \newtext{structure-based} methods for visual localization, which do not critically rely on storing visual descriptors, \hl{thereby reducing storage, relieving privacy concerns and eliminating the need for descriptor maintenance.}

\section{Related Work}\label{sec:related_work}
\PAR{Structure-Based Localization.}
\camr{Methods of this kind}
\cite{Sattler2017PAMI,Sarlin2019CVPR,Taira2018CVPR,Schoenberger2018CVPR,Brachmann2017CVPR} \newtext{commonly establish explicit correspondences} between the query image pixels and the 3D points of the environment to compute the query image pose from the established matches using PnP solvers~\cite{Kneip2011CVPR,Gao2003TPAMI}. 
\camr{Keypoint correspondences are made}
by computing and matching visual descriptors for each keypoint from a query and database images~\cite{Dusmanu2019CVPR,Detone2018CVPRW,Wang2020ECCV,Luo2020CVPR,Germain2020ECCV,Sarlin2020CVPR, Sun2020CVPR}.
Another recent work~\cite{Sarlin2021CVPR} iteratively optimizes a camera pose by minimizing visual descriptor distances between the 3D points observed in the query and the reference images.
\newtext{While it does not establish matches, it relies on visual descriptors extracted from a neural network and requires 3D points.}
Structure-based localization methods achieve impressive localization accuracy and state-of-the-art performance~\cite{Sarlin2019CVPR,Detone2018CVPRW,Sarlin2020CVPR} in the long-term localization benchmark~\cite{Sattler2018CVPR,Toft2020TPAMI}.

\PAR{Practical Challenges in Structure-Based Localization.}
\label{sec:exp-challenges}
\begin{table}[t]
\caption{On the challenges of large-scale structure-based localization. 
\camr{Analysis is performed on the MegaDepth~\cite{Li2018CVPR} composed of many landmarks (similar to city districts), acting as an example of a city-scale dataset.}
We compare visual-based matching (VM) and geometric-based matching (GM) methods by analysing their storage requirement and considering whether a method \camr{requires to maintain map descriptors} as well as provides privacy protection (\cf the supplementary for more details.)
\camr{For structured-based localization, scene coordinates (3D) and camera metadata (Cameras) are stored to obtain 2D-3D correspondences.
In contrast to VM methods that need to additionally store visual descriptors or extract descriptors on-the-fly from the raw images, we show that using GM instead of VM, significantly reduces storage requirements,
safeguards user privacy and bypasses the need for descriptor maintenance~\cite{Dusmanu2021ICCV}.}
}
\label{tab:storage_plans}
\centering
\setlength{\tabcolsep}{3pt}
\resizebox{\textwidth}{!} {
\begin{tabular}{l| l | c c |c c c c c }
\toprule
&\multirow{3}{*}{Method}  & Desc. & Privacy & \multicolumn{4}{c}{Database Storage (GB, $\downarrow$)} \\ 
& & Maintenance & & Cameras (MB) & 3D & Raw Ims & Descs & Total \\
\midrule
\multirow{3}*{VM} 
& SIFT~\cite{Lowe2004IJCV}           &\xmark  &\xmark & 15.73 &3.44 & \xmark & 130.10 (uint8) &  133.33  \\
& SuperPoint~\cite{Detone2018CVPRW}     &\xmark  &\xmark & 15.73  &3.44 & \xmark & 1040.76 (fp32) & 1044.21 \\
& Extract on-the-fly &\xmark  &\xmark & 15.73  &3.44 & 157.84 & \xmark & 161.29 \\
\midrule
\multicolumn{2}{c|}{Geometric-based Matching}
&\cmark  &\cmark  &   15.73 & 3.44 & \xmark & \xmark & \textbf{3.45} \\
\bottomrule
\end{tabular}
}
\end{table}
\camr{Despite being highly accurate, modern localization solutions encounter practical challenges when deployed onto real-life applications, spanning city-level scale. The challenges are threefold: i)
Relying on visual descriptors~\cite{Detone2018CVPRW,Dusmanu2019CVPR,Wang2020ECCV,Luo2020CVPR} makes the system demanding in storage\footnote{\emph{Storage} as in non-volatile preservation of data, in contrast to volatile \emph{memory}.}
as shown in \cref{tab:storage_plans}. 
}
\camr{To reduce storage requirement of the 3D scene representation, compression can be done by keeping a subset of the 3D points~\cite{Cao2014CVPR, Camposeco2019CVPR, MeraTrujillo20203DV} and quantising~\cite{Camposeco2019CVPR, TranTIP2019,Cheng2019ICCV} the descriptors associated with the 3D points.
HybridSC~\cite{Camposeco2019CVPR} stands out among
the existing work, with its extreme compression rate
and minimal accuracy loss.
}
\camr{ii) Localization methods following a server-client model need to transmit visual descriptors between the server and client}, which exposes the model to a risk of a privacy breach~\cite{Pittaluga2019CVPR,Dosovitskiy2016CVPR,Dosovitskiy2016NeurIPS,Chelani2021CVPR}. 
To mitigate this issue, recent work~\cite{Ng2022CVPR, Dusmanu2020CVPR} developed descriptors that are more robust against privacy attacks with slightly lower accuracy.
\camr{iii) With the ongoing advancements in local features methods~\cite{Detone2018CVPRW,Dusmanu2019CVPR,Wang2020ECCV,Luo2020CVPR, Ng2022CVPR, Dusmanu2020CVPR}, continuously updating scene descriptors is a foreseeable demand~\cite{Dusmanu2021ICCV} for visual-based matching methods.
However, such an update requires either re-building the map with new descriptors or transforming the existing descriptors~\cite{Dusmanu2021ICCV} to new ones.
}
In this paper, we propose an \textit{orthogonal} direction to address the storage, privacy and  descriptor maintenance challenges in structure-based localization by relying solely on more \newtext{lightweight} geometric information for matching.

\PAR{End-to-End Learned Localization.} 
A recent trend of methods leverage data-driven techniques to learn to localize in an end-to-end manner, without relying on point clouds. This is achieved by either regressing scene coordinates, regressing the camera's absolute pose or regressing its relative pose \wrt to a database image.
Scene coordinate regression methods~\cite{Brachmann2017CVPR,Brachmann2018CVPR,Brachmann2019ICCV,Bhowmik2020CVPR,Cavallari20193DV,Li2020CVPR,Yang2019CVPR} directly regress dense 3D scene coordinates from 2D images. 
However, they need to be re-trained for every new scene due to their lack of generalization
~\cite{Brachmann2017CVPR,Brachmann2018CVPR,Cavallari20193DV,Brachmann2019CVPR}.
In certain cases, multiple instances of the same network are trained on sub-regions of the scene, due to the limited capacity of a single network~\cite{Brachmann2019CVPR}.
Therefore, it is unclear how to scale these methods~\cite{Brachmann2017CVPR,Brachmann2018CVPR,Brachmann2019ICCV,Bhowmik2020CVPR,Cavallari20193DV,Li2020CVPR,Yang2019CVPR}, that are traditionally evaluated only on small indoor rooms, to large-scale scenes.
Absolute pose regression \camr{(APR)} methods implicitly encode the scene representation inside the network
and directly regress the pose from the query image~\cite{Kendall2015ICCV,Kendall2016ICRA,Kendall2017CVPR,Radwan2018RAL,Walch2017ICCV}.
\camr{While earlier} methods required training a model per scene and have been shown to overfit to the viewpoints and appearance of the training images~\cite{Sattler2019CVPR},
\camr{recent work in multi-scene APR \cite{Blanton2020CVPRW,Shavit2021ICCV}
loosened the per-scene training requirements.}
\camr{Compared to multi-scene APR,
our method generalizes
across scenes as other structure-based localization methods (\cf \cref{sec:exp-generalization}) while addressing its aforementioned practical challenges.}
Another related approach that sidesteps maintaining a 3D model with visual descriptors, is to regress relative camera poses~\cite{Ding2019ICCV,Zhou2020ICRA,Balntas2018ECCV,Laskar2017ICCVW} from a query image to its relevant database images.
However, directly regressing the geometric transformations in general leads to limited generalization~\cite{Sattler2019CVPR,Zhou2020ICRA}.

\PAR{Direct Geometric Keypoint Matching.}
\camr{Matching image keypoints directly to 3D point clouds
while jointly estimating pose has been widely investigated 
under relatively constrained environments \cite{David2004IJCV,MorenoNoguerECCV2008,Brown2015ICCV,Campbell2017ICCV,Campbell2019CVPR,Liu2020arXiv,Campbell2020ECCV}. Some require pose initialization~\cite{David2004IJCV} or  pose distribution priors~\cite{MorenoNoguerECCV2008}, while others, based on globally optimal estimators, have prohibitive runtime requirements in order to produce accurate estimates \cite{Brown2015ICCV,Campbell2017ICCV,Campbell2019CVPR}.}
\camr{In contrast, the recent state-of-the-art, data-driven, geometric matching approaches \cite{Liu2020arXiv,Campbell2020ECCV} strike a good compromise between pose accuracy and time required to produce an accurate estimate. Despite not producing globally optimal solutions, BPnPNet~\cite{Campbell2020ECCV} is able to estimate a reliable pose in a fraction of a second.}
Given a set of 2D keypoints in the query image and a set of 3D points in the scene point cloud, BPnPNet jointly estimates matches between these two sets \textit{purely} based on geometric information. 
However, this approach was shown to work in idealistic scenarios assuming no outlier keypoints
and, as we experimentally demonstrate, the matching performance degrades significantly once outliers are introduced. 
The outlier-free assumption clearly does not hold for challenging real-world localization scenarios as map building and keypoint detection are all challenging tasks, prone to errors and noise. In our work, we build upon BPnPNet and
design a geometric matching module that is robust against keypoint outliers. We show  in \cref{sec:exp-ablations} that our approach significantly outperforms BPnPNet in matching keypoints with noisy outliers, effectively enabling the applicability of geometric-based matching to real-world visual localization.

\begin{figure}[t]
    \centering
    \includegraphics[width=0.87\textwidth]{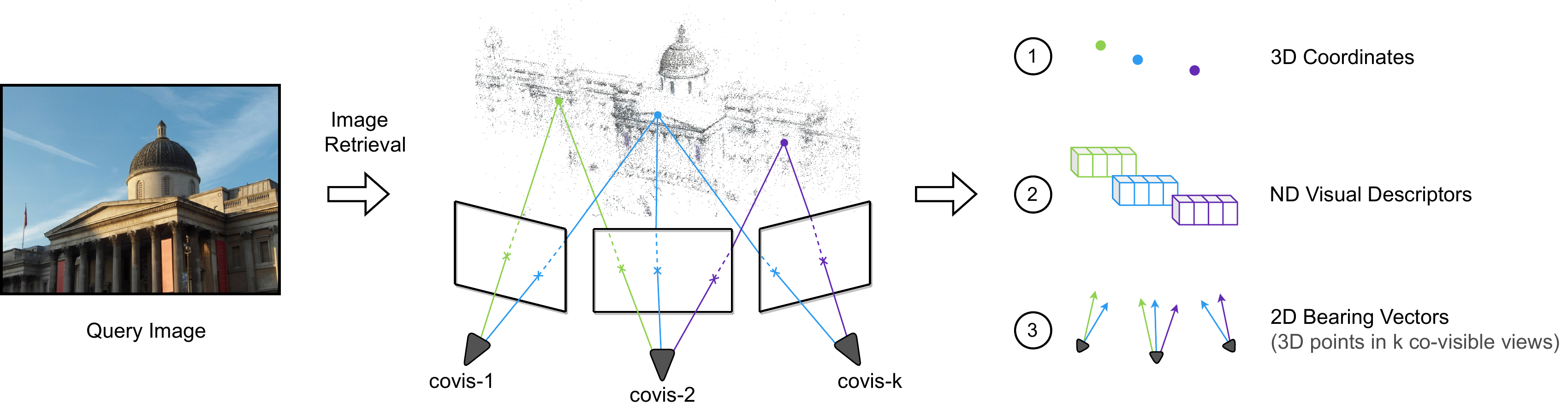}
    \caption{Co-visible views \& keypoint representations. \newtext{Retrieving co-visible reference images (views) of a query image, narrows the matching against a full 3D point cloud to a subset of points that are more likely to be visible to the query image. Each 3D point can be represented differently by:} 1) its 3D coordinate; 2) a visual descriptor that incorporates local appearance; or 3) a bearing vector that represents the direction from the reference camera origin to a 3D point in normalized coordinates. In this paper, we explore keypoint matching using representations 1) and 3).
    }
    \label{fig:covisibility-representations}
\end{figure}

\section{Task Definition}
\label{sec:task_def}

\PAR{Structure-Based Localization Pipeline.}
Structure-based methods assume as input a query image, a 3D point cloud of the scene, \newtext{and database images with known poses.}
These methods first retrieve a set of database images that are co-visible with the query image, \ie, have a visual overlap, as illustrated in \cref{fig:covisibility-representations}. 
Next, after narrowing down the search space, they establish 2D-3D correspondences between the query image keypoints and a (retrieved) subset of the 3D point cloud. %
This set of correspondences can be used to estimate the query image pose using a PnP solver~\cite{Ke2017CVPR,Gao2003TPAMI}. 
The majority of prior work~\cite{Dusmanu2019CVPR,Detone2018CVPRW,Wang2020ECCV,Luo2020CVPR,Germain2020ECCV,Sarlin2020CVPR,Sarlin2019CVPR} rely on storage-consuming visual descriptors, stored together with the point cloud, to establish 2D-3D matches. %
The key challenge we address is how to establish \newtext{those} correspondences 
\emph{without} visual descriptors. 

\PAR{Problem Formulation.}
We assume two point sets, one with 2D keypoint coordinates in the image plane $\vec p_i \in \R^2$, and the second containing 3D point coordinates $\vec q_j \in \R^3$. We seek the matching set $\mathcal{M} := \{(i, j) | \vec p_i = \pi(\vec q_j; \mat K, \mat R, \vec t)\}$,
\ie, the set of index pairs $i$ and $j$, for which if the $j$-th 3D keypoint is projected to the image plane, it matches the coordinates specified by the corresponding $i$-th 2D point.
The camera intrinsic matrix $\mat K \in \R^{3 \times 3}$ is assumed to be known, and the operator $\pi(\cdot)$ represents the camera projection function, which transforms 3D points onto the camera's frame of reference and projects them to the image plane according to the camera's intrinsics. 
Our goal is to find the correct 2D-3D keypoint matches for accurate pose estimation.

\PAR{Keypoint Representation.} 
We represent 2D pixels using 2D coordinates $(u, v) \in \mathbb{R}^2$ in the image plane. 
To learn a matching function that is agnostic to different camera models, we 
uplift those 2D points into a bearing vector representation $\vec b \in \R^2$\newtext{, effectively removing the effect of the camera intrinsics}.
Bearing vectors encode the direction (or bearing) of points in a camera's frame of reference.
We compute bearing vectors from image pixels as: $[\vec b^\top \, 1]^\top \propto \mat K^{-1} [u \, v \,1]^\top$.
For a 3D point, we consider two different representations (see \cref{fig:covisibility-representations}):
(i) as 3D coordinates $(x, y, z) \in \mathbb{R}^3$ \wrt a 3D world reference/origin; \newtext{and}
(ii) as a bearing vector \wrt a reference database image.
The bearing vector representation allows bringing both 2D pixels and 3D points to the same data modality.
Given a 3D point $\vec p \in \R^3$ and transformation ($\mat R$, $\vec t$) from the world to the database image's frame of reference, we compute the corresponding bearing vector as:
\begin{equation}
    \vec p' = \mat R \vec p + \vec t, \,\,\, [\vec b^\top \, 1]^\top =  \vec p' / {p'_z},
\end{equation}
\newtext{where $\vec p'$ represents $\vec p$ in the camera's frame of reference, and $\vec p'_z$ represents its $z$ coordinate}.
\newtext{As shown in \cref{tab:storage_plans}, these geometric-based point representations require significantly lower storage compared to visual descriptor based ones, \eg, as low as 3\% compared to the storage of modern descriptors.
}

\section{Geometric-Only Matching}
\PAR{BPnPNet in a Nutshell.}
BPnPNet~\cite{Campbell2020ECCV}
made great progress
towards establishing correspondence between the query keypoints and 3D point cloud in the absence of visual descriptors.
\newtext{It} proposes an end-to-end trainable, differentiable matcher that performs 2D \newtext{to} 3D cross modal matching without relying on appearance information.
While this is a step in the right direction, we 
\newtext{show in ~\cref{sec:exp-ablations}} that it does not scale to the real-world visual localization scenarios where outliers, \ie points without a match, are pervasive. Direct 2D-3D matching of sparse keypoints is a challenging problem due to low amount of discriminative data, \ie points no longer have a local visual appearance, and its cross-modal nature.
In a nutshell, BPnPNet (i) encodes points to obtain per point features, 
(ii) establishes matches using the Sinkhorn algorithm~\cite{Cuturi2013NIPS,Sinkhorn1967PJM}, 
\newtext{which finds the optimal assignment between geometrical features,} and finally, (iii) leverages a differentiable PnP solver that imposes an additional pose supervision on the network. 
In the following, we build on the observation that the lightweight geometric feature encoder does not possess the necessary representational power to produce features that generalize simultaneously to situations with and without outliers. 

\begin{figure}[t]
    \centering
    \includegraphics[width=\textwidth]{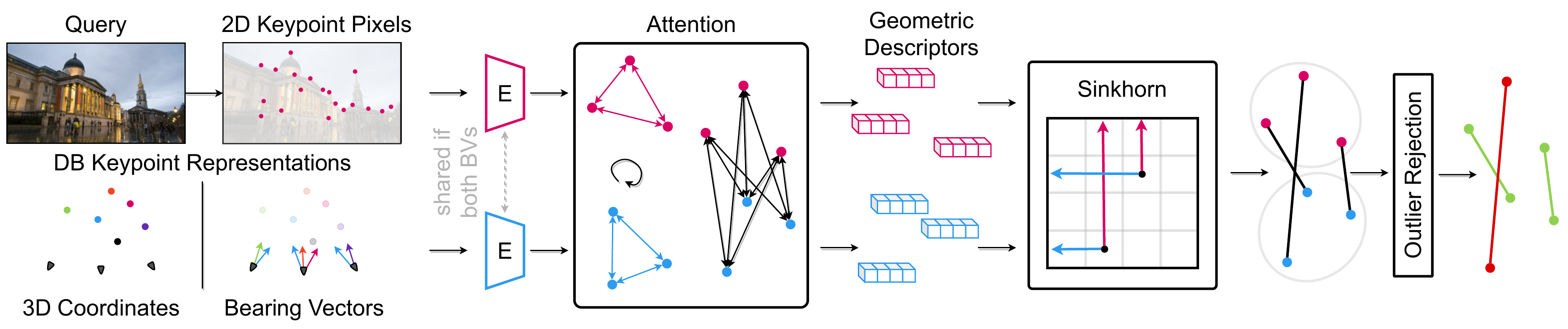}
    \caption{\newtext{GoMatch components overview.} The query image and database keypoints first undergo a feature encoder \textbf{E} to generate per-point features. \newtext{We share encoders in the query and database branch when database points are represented as bearing vectors otherwise not.}
    These features are refined in the attention layer and \newtext{then used} in the Sinkhorn matching stage \newtext{to} establish an initial set of candidate matches\newtext{, from which erroneous matches are filtered with an outlier rejection layer. }
    }
    \label{fig:overview}
\end{figure}
\subsection{GoMatch: Embracing Outliers}
In GoMatch we (i) propose architectural changes that enable resilience to outliers and  (ii) cast the \emph{cross-modal} nature of 2D-3D matching to an \emph{intra-modal} setting through the use of bearing vectors. 
Below, we explain the details of these contributions, 
\newtext{which are experimentally validated to be all necessary and critical to outlier-robust geometric matching in \cref{sec:exp-ablations}.}
We refer to \cref{fig:overview} for a visual overview of the entire network.
Furthermore, we add an outlier rejection layer 
to retain only quality matches from the Sinkhorn outputs. 
While we introduce the novel network components in the following paragraphs, we refer the reader to the supplementary material for an in-depth description of all network components.

\PAR{Feature Refinement through Attention.} In BPnPNet, each keypoint node is processed in parallel with an MLP-style encoder to extract features directly for matching, and information exchange happens only in the Sinkhorn matching stage. 
This might lead to a learned feature representation which lacks context information within each 2D/3D modality and cross modality. 
Based on this assumption, we explore adding information exchange prior to matching.
To enhance the context information within each modality, we apply \textit{self-attention} to the raw encoded features where a graph neural network~\cite{Huang2021CVPR} refines features of every keypoint by exchanging the information with a fixed number of closest neighbors in coordinate space.
This is followed by \textit{cross-attention}~\cite{VaswaniNeurIPS2017}, where every keypoint from one modality will interact with all keypoints from the other modality through a sequence of multi-head attention layers.
By stacking several blocks of such self-/cross-attention layers, we are able to learn more representative features, which allows Sinkhorn to identify significantly better outlier matches.

\PAR{Outlier Rejection.}
After Sinkhorn matching, the estimated corresponding pairs may still contain outlier matches. 
To filter those,
we follow \cite{Moo2018CVPR} and add a classifier that takes in the concatenated geometric features from the query and database keypoints, and predicts confidence scores for all matches. Estimated correspondences with confidence below a threshold ($0.5$ in practice) are rejected. 

\PAR{Matching with Bearing Vectors.} 
\newtext{Directly matching 2D keypoints to cross-modal 3D coordinates} is challenging because it requires the network to learn features that have to consider not only the relationship between keypoints, but also the influence of different camera poses.
Furthermore, the different distributions of 3D point clouds between datasets, \eg, different scene sizes or different gravity directions, are particularly challenging for a single encoder to learn. 
Based on this observation, we propose to leverage the \emph{bearing vector} representation of the database points to 
\camr{sidestep the difference in data modalities.}
\camr{In addition to nullifying the effects of the camera intrinsics, projecting 3D points as bearing vectors onto a “covisible” frame that is closer to the query frame (compared to the world reference frame), effectively mitigates the influence of the camera pose (viewpoint changes) during matching, albeit dependent on the quality of retrieval.}
Finally, bearing vectors provide a common modality between query and database keypoints, eliminating the need for a separate encoder. 
As we demonstrate in our experimental section, the change in input type has a substantial positive effect.

\subsection{Training GoMatch} 
All of our models are trained to learn feature matching and outlier filtering jointly, using a matching loss and an outlier rejection loss.

\PAR{Matching Loss.}
The Sinkhorn matching layer is trained to output a discrete joint probability distribution of two sets of keypoints being matched. We denote this distribution as $\tilde{\mat P} \in \R^{M+1 \times N+1}_+$, such that $\sum_{i=1}^{M+1} \sum_{j=1}^{N+1} \tilde{\mat P}_{ij} = 1$, \ie, is a valid probability distribution. 
Here, $M$ and $N$ denote the total number of query and database keypoints considered during the matching. We include an extra row and column to allow keypoints not to be matched. 
We employ a negative log loss to the joint discrete probability distribution. Consider the set of all ground truth matches $\mathcal{M}$, as well as the set of unmatched query keypoints $\mathcal{U}_{q}$ and database keypoints $\mathcal{U}_{d}$. The matching loss is of the form:
\begin{equation}
    L_{\text{match}} = - \frac{1}{N_m} \biggl(\sum_{(i, j) \in \mathcal{M}} \log \tilde{\mat P}_{ij} +  \sum_{i \in \mathcal{U}_{q}} \log \tilde{\mat P}_{i(N + 1)} + \sum_{j \in \mathcal{U}_{d}} \log \tilde{\mat P}_{(M + 1)j} \biggr),
\end{equation}
where $N_m = |\mathcal{M}| + |\mathcal{U}_{q}| + |\mathcal{U}_{d}|$.

\PAR{Outlier Rejection Loss.}
For the outlier rejection layer we employ a mean weighted binary cross-entropy loss:
\begin{equation}
    L_{\text{or}} = - \frac{1}{N_c}\sum_{i=1}^{N_c} w_i \left(y_i \log p_i + (1 - y_i) \log (1 - p_i) \right),
\end{equation}
where $N_c$ denotes the total number of correspondences supplied to the outlier rejection layer. The term $p_i$ denotes the classifier output probability for each correspondence, while $y_i$ denotes the correspondence target label, and $w_i$ is the weight balancing the negative and positive samples. Our final loss balances both terms equally, \ie, $L_{\text{total}} = L_{\text{match}} + L_{\text{or}}$.
We present implementation details about training and testing process in our supplementary material.
\section{Experimental Evaluation}\label{sec:experiments}
\newtext{In this section, we thoroughly study the potential of using our proposed geometric-based matching for the task of real-world visual localization. We start our experiments by testing the robustness of BPnPNet~\cite{Campbell2020ECCV} and GoMatch with keypoint outliers. Next, we verify our technical contribution of successfully diagnosing the missing components leading to robust geometric matching and enabling geometry-based visual localization.
Furthermore, we position gemetric-based localization among other state-of-the-art visual localization approaches by comprehensively analysing each method in terms of localization accuracy, descriptor maintenance effort~\cite{Dusmanu2021ICCV}, privacy risk, and storage demands (\cref{sec:exp-compar_sota}).
Finally, we present a generalization study (\cref{sec:exp-generalization}) to highlight that our proposed method generalizes across different types of datasets and keypoint detectors.
We hope that our in-depth study serves as a starting point of this rarely explored new direction, and inspires new work to advance scalable visual localization through geometric-only matching in the future. 
}

\subsection{Datasets}\label{sec:exp-datasets}

\newtext{
We use MegaDepth~\cite{Li2018CVPR} for training and ablations, given its large scale. It consists of images captured in-the-wild from 196 outdoor landmarks.
We adopt the original test set proposed in~\cite{Li2018CVPR}, and split the remaining sequences into training and validation sets. 
After verifying our best models on Megadepth, we evaluate them on the popular Cambridge Landmarks~\cite{Kendall2015ICCV} (Cambridge) dataset which consists of 4 outdoor scenes of different scales.
It allows for convenient comparison to other localization approaches. We use the reconstructions released by~\cite{Sarlin2021CVPR}.
In addition, we evaluate on the indoor 7-Scenes~\cite{Shotton2013CVPR} dataset to further assess the generalization capability of our method. 7-Scenes is composed of dense point clouds captured by an RGB-D sensor, and thus provides an alternative environment with different keypoint distributions, in both 2D images and 3D point clouds.
We perform evaluation on the official test splits released by the Cambridge and 7-Scenes datasets.
We provide detailed information about training data generation using MegaDepth in the supplementary.
}

\subsection{Experimental Setup}
\PAR{Keypoint Detection.} 
\newtext{
For MegaDepth and Cambridge, we use respectively SIFT~\cite{Lowe2004IJCV} and SuperPoint~\cite{Detone2018CVPRW}, preserving the same keypoint detector
used to reconstruct their 3D models.
For 7-Scenes, we use both SIFT and SuperPoint to extract keypoints for both 2D images and 3D point cloud given RGB-D images.}

\PAR{Retrieval Pairs.}
\newtext{
We use ground truth to sample retrieval pairs that have at least 35\% visual overlap in MegaDepth
to ensure enough matches are present during training, as well as to isolate the side-effect of retrieval performance during ablations.
For evaluation and comparison to state-of-the-art localization methods, we follow~\cite{Sarlin2021CVPR} and use their \textit{top-10} pairs retrieved using NetVLAD~\cite{Arandjelovic2016CVPR} on Cambridge and DenseVLAD~\cite{Torii2015CVPR} on 7-Scenes.
}

\PAR{Matching Baselines.}  
\newtext{
We consider BPnPNet~\cite{Campbell2020ECCV} as our geometric-based matching baseline.
For a fair comparison, we re-train BPnPNet
using our training data. 
Our visual-based matching baselines use SIFT~\cite{Lowe2004IJCV} and SuperPoint~\cite{Detone2018CVPRW} \camr{(SP)} as keypoint descriptors.
To match visual descriptors, we use  nearest neighbor search~\cite{Muja2014PAMI} with mutual consistency %
\camr{by default} 
\camr{and SuperGlue~\cite{Sarlin2020CVPR} (SG)}.
}

\PAR{Localization Pipeline.}
\camr{Following the state-of-the-art structure-based localization, \eg, HLoc~\cite{Sarlin2019CVPR}}, we first obtain up to $k=10$ retrieval pairs between a query and database images. Then we establish per-pair 2D to 3D matches using either a geometric-based or a visual-based matching model, and then merge results from $k$ pairs based on their matching scores to estimate camera poses. For fairness, all matching baselines use identical retrieval pairs and identical settings for the PnP+RANSAC solver~\cite{Ke2017CVPR}.

\PAR{Evaluation Metrics.}
\newtext{
For MegaDepth, we follow BPnPNet~\cite{Campbell2020ECCV} to report the pose error quantiles at 25/50/75\% for the translation and rotation ($^{\circ}$) errors as evaluation metrics.
However, as the scale unit of MegaDepth
is undetermined
and varies between scenes,
the translation errors are
not consistent between scenes.
Therefore, we propose a new metric based on pixel-level reprojection errors
that preserves scene consistency.
For each query, we project its inlier 3D keypoints using the predicted and the ground-truth poses. We then report the area under the cumulative curve (AUC) of the mean reprojection error up to $1/5/10$px, inspired by the pose error based AUC metric used in \cite{Sun2021CVPR, Sarlin2021CVPR}.
We report the commonly used median translation ($m$) and rotation ($^{\circ}$) errors~\cite{Kendall2015ICCV, Sattler2019CVPR, Camposeco2019CVPR} per-scene on Cambridge and 7-Scenes. 
}

\subsection{Ablations}
\label{sec:exp-ablations}
\newtext{We perform ablation studies with MegaDepth's~\cite{Li2018CVPR} test split, where all retrieval pairs have guaranteed 35\% co-visibility, to focus purely on matching performance. In addition, we study the effect of using a single co-visible reference view ($k=1$) as a minimal setting, as well as multiple views, \eg, $k=10$, following the common practice in hierarchical structure-based localization~\cite{Sarlin2021CVPR, Sattler2019CVPR}. 
To better understand the new AUC metric, we also present an \textbf{Oracle} that uses ground truth matches as its prediction. It is used to show the upper-bound performance that can be achieved using our metric and generated data.}

\begin{figure}[!t]
    \centering
    \includegraphics[width=0.8\textwidth]{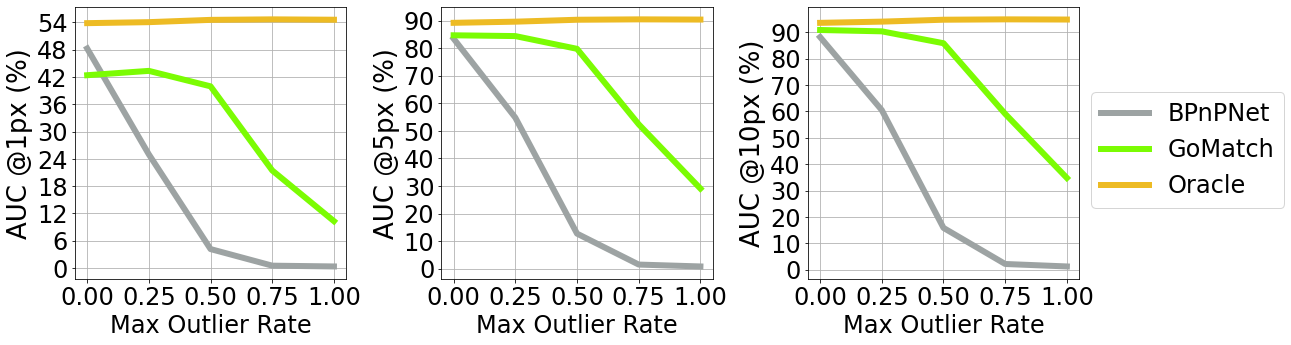}
    \caption{Influence of keypoint outlier rate. \newtext{In contrast to prior work BPnPnet~\cite{Campbell2020ECCV}, GoMatch is significantly more robust against keypoint outliers thanks to the more powerful attention-based architecture as well as our novel formulation of matching bearing vectors instead of cross-modal features.}}
    \label{fig:orate_abalt}
\end{figure}

\PAR{Sensitivity to Keypoint Outliers.} In a real-world localization setting, the detected query image keypoints will often be noisy and will not have a direct correspondence in the 3D point cloud. Keypoint matching methods thus need to be able to cope with outliers. %
We first study whether our baseline has this capability by manually increasing the maximum outlier rate, ranging from 0 to 1.
\camr{The outlier rate is computed as the number of keypoints without a match divided by the total number of keypoints, taking the maximum between 2D and 3D.}
\newtext{For all other experiments, we do not control keypoint the outlier rate to properly mimic realistic conditions.}
\newtext{As shown in \cref{fig:orate_abalt}, the Oracle stays round 55/90/94\% (AUC@1/5/10px).
The large error at 1px is due to our match generation process
(\cf supplementary for a detailed discussion).
BPnPNet~\cite{Campbell2020ECCV} slightly outperforms GoMatch at 1px threshold, being similarly accurate to us at 5/10px thresholds in the absence of outliers.
However, as the ratio of outliers increases, the performance of BPnPNet drastically drops,
while GoMatch gracefully handles outliers,
\ie, GoMatch is always above 80\% at 5/10px up to 50\% of outliers.
This experiment confirms that GoMatch is significantly more robust to outliers compared to BPnPNet. 
This outlier robustness is achieved through careful modifications to the network architecture and 3D point representation, both validated by a thorough performance analysis presented in the next sections.
}

\begin{table}[t]
\caption{\newtext{GoMatch ablation. \textit{Top:} We present Oracle for reference and re-trained BPnPNet~\cite{Campbell2020ECCV} as our baseline. \textit{Middle:} We study how the 3D representation (Repr.) and architectural changes influences the performance. Using bearing vector (BVs) instead of 3D coordinates (Coords) as representation and introducing feature attention (Att) are the most crucial factors to the performance improvement. Together with further benefits from the outlier rejection (OR) component and sharing the query and database keypoint feature encoders leads us to the full GoMatch model (\textit{Bottom}). All results rely on a singe retrieval image unless stated otherwise, \eg, $k=10$.}}
\label{tab:ablat_archs}
\centering
\setlength{\tabcolsep}{3pt}
\resizebox{12cm}{!} {
\begin{tabular}{l c c c c c c c}
\toprule
\multirow{2}{*}{Model}& 3D & Share & \multirow{2}{*}{Att} & \multirow{2}{*}{OR} & Rotation ($^{\circ}$) &  Translation & Reproj. AUC (\%) \\ 
 & Repr. & Encoder &  &  &\multicolumn{2}{c}{Quantile@25/50/75\% ($\downarrow$)} & $@1/5/10$px ($\uparrow$) \\ 
\midrule
Oracle  &  &  & &  & 0.03/0.06/0.10 & 0.00/0.00/0.01 & 54.58/90.37/94.87 \\
BPnPNet & Coords & \xmark & \xmark & \xmark & 15.17/31.05/59.78 & 1.67/3.14/5.31 & 0.34/0.83/1.21 \\
BPnPNet ($k=10$) & Coords & \xmark & \xmark & \xmark & 16.03/33.27/63.90 & 1.59/3.24/5.80 & 0.56/1.08/1.50 \\
\midrule
\multirow{6}{*}{Variants} 
  & BVs & \xmark & \xmark & \xmark & 12.19/27.68/58.22 & 1.26/2.8/5.14 & 0.37/1.48/2.18 \\  
  & BVs & \cmark & \xmark & \xmark & 9.16/22.62/53.20 & 0.98/2.38/4.72 & 0.85/3.09/4.36\\ 
  & BVs & \cmark & \cmark & \xmark & 0.55/8.08/29.34 & 0.05/0.84/3.34 & 9.13/25.71/31.65  \\  
  & BVs & \xmark & \cmark & \cmark & 0.38/7.46/31.75 & 0.04/0.83/3.73 & 10.22/28.17/33.69\\ 
  & \cellcolor{cgrey}Coords & \cellcolor{cgrey}\xmark & \cellcolor{cgrey}\cmark & \cellcolor{cgrey}\cmark & \cellcolor{cgrey}4.09/23.56/63.21 & \cellcolor{cgrey}0.37/2.53/5.93 & \cellcolor{cgrey}3.81/13.54/17.46\\   
\midrule  
GoMatch & \cellcolor{cgreen}BVs & \cellcolor{cgreen}\cmark & \cellcolor{cgreen}\cmark & \cellcolor{cgreen}\cmark & \cellcolor{cgreen}0.36/6.97/29.85 & \cellcolor{cgreen}0.03/0.69/3.38 & \cellcolor{cgreen}10.30/29.08/34.79 \\
GoMatch($k=10$) & \cellcolor{cgreen}BVs & \cellcolor{cgreen}\cmark & \cellcolor{cgreen}\cmark & \cellcolor{cgreen}\cmark & \cellcolor{cgreen}\textbf{0.15/0.95/13.00} & \cellcolor{cgreen}\textbf{0.01/0.09/1.55} & \cellcolor{cgreen}\textbf{15.14/42.39/51.24} \\
\bottomrule
\end{tabular}
}
\end{table}

\PAR{Architecture-Level Analysis.}  \newtext{
In \cref{tab:ablat_archs} (\textit{Top}), we present the Oracle and BPnPNet~\cite{Campbell2020ECCV} re-trained on our data for a direct comparison with GoMatch.
This is paired with additional variants, progressively transitioning from BPnPNet to GoMatch. %
We found that shared encoding brings performance gains up to 0.48/1.61/2.18 AUC percentage points. 
Adding feature attention on top leads to a significant improvement of 8.28/22.62/27.29 AUC percentage points.
By further adding the outlier rejection increases the AUC by 1.17/3.37/3.14 percentage points.
We conclude that these network components yield 9.93/27.6/32.61 percentage points of improvements in terms of AUC scores when using bearing vectors the representation.
}

\PAR{Representation-Level Analysis.} \newtext{
Using 3D coordinates (Coords) instead of bearing vectors (BVs), even with attention and outlier rejection, hinders performance dramatically by 6.49/15.54/17.33 percentage points. 
If we only change the representation from Coords to BVs,  without attention nor outlier rejection, the improvement is merely 0.31/1.29/1.9 percentage points.
Therefore, we verify the bearing vector representation is as important as the architectural changes,
and both contribute towards keypoint outlier resilience.
By modifying both architecture and representation, GoMatch outperforms the re-trained BPnPNet by 9.96/28.25/33.58 AUC percentage points.
}

\PAR{Utilizing Multiple Co-visible Images.}  \newtext{As shown in \cref{tab:ablat_archs}, when using $k=10$ co-visible views, both methods improved their result:  BPnPNet by a small margin and GoMatch by a large margin of 4.84/13.31/16.45 AUC percentage points. 
We thus use $k=10$ for all of the following experiments.
}

\subsection{Comparison to Localization Baselines}\label{sec:exp-compar_sota}

\begin{table}[t]
\caption{Comparison to existing localization baselines. We consider end-to-end (E2E) methods and structure-based methods that either matches visual descriptors (VM) or geometries (GM). \camr{We report median translation and angular error for each landmark and combined storage requirements for operating on all landmarks.} 
\textit{No Desc. Maint.} is checked if a method does not require %
descriptor updates in the long run.
\camr{\textit{Privacy} is checked if a method is resilient to existing known privacy attacks.}}
\label{tab:compare_sota_cambridge}
\centering
\setlength{\tabcolsep}{3pt}
\resizebox{\textwidth}{!} {
\begin{tabular}{l l c c c c c c c }
\toprule
&\multirow{2}{*}{Method} & Storage & No Desc. & \multirow{2}{*}{Privacy} & King’s College & Old Hospital &  Shop Facade & St. Mary’s Church \\ 
& &  (MB)  & Maint. &  & \multicolumn{4}{c}{Median Pose Error (m, $^{\circ}$) ($\downarrow$)}\\
\midrule
\multirow{3}*{\begin{sideways} E2E \end{sideways}} 
& PoseNet~\cite{Kendall2015ICCV}    & 200 & \camr{\cmark} & \cmark & 1.92/5.40 & 2.31/5.38 & 1.46/8.08 & 2.65/8.48 \\
& DSAC++~\cite{Brachmann2018CVPR}   & 828 & \camr{\cmark} & \cmark & 0.18/0.30 & 0.20/0.30 & 0.06/0.30 & 0.13/0.40 \\
& \camr{MSPN~\cite{Blanton2020CVPRW}}   & \camr{-} & \camr{\cmark} & \camr{\cmark} & \camr{1.73/3.65} & \camr{2.55/4.05} & \camr{2.92/7.49} & \camr{2.67/6.18} \\
& \camr{MS-Transformer~\cite{Shavit2021ICCV}}   & \camr{71.1} & \camr{\cmark} & \camr{\cmark} & \camr{0.83/1.47} & \camr{1.81/2.39} & \camr{0.86/3.07} & \camr{1.62/3.99} \\
\midrule
\multirow{3}*{\begin{sideways} VM \end{sideways}} 
& HybridSC~\cite{Camposeco2019CVPR} & 3.13 & \xmark & \textbf{\textit{$?$}} & 0.81/0.59 & 0.75/1.01 & 0.19/0.54 & 0.50/0.49 \\
& Active Search~\cite{Sattler2017PAMI} & 812.7   & \xmark & \xmark & 0.42/0.55 & 0.44/1.01 & 0.12/0.40 & 0.19/0.54 \\
& \camr{HLoc~\cite{Sarlin2019CVPR}(w.SP~\cite{Detone2018CVPRW})} & 3214.84  &\xmark  &\xmark  & 0.16/0.38 & 0.33/1.04 & 0.07/0.54 & 0.16/0.54 \\
& \camr{HLoc(w.SP+SG~\cite{Sarlin2020CVPR})} & 3214.84  &\xmark  &\xmark  & \textbf{0.12/0.20} & \textbf{0.15/0.30} & \textbf{0.04/0.20} & \textbf{0.07/0.21} \\
\midrule
\multirow{2}*{\begin{sideways} GM \end{sideways}} 
& BPnPNet~\cite{Campbell2020ECCV}   & 48.15 & \cmark & \cmark & 26.73/106.99 & 24.8/162.99 & 7.53/107.17 & 11.11/49.74 \\
& GoMatch                        & 48.15 & \cmark & \cmark & 0.25/0.64 & 2.83/8.14 & 0.48/4.77 &  3.35/9.94 \\
\bottomrule
\end{tabular}
}
\end{table}

Following the discussion in~\cref{sec:exp-challenges}, we comprehensively compare GoMatch with other established baselines by looking beyond localization performance, and considering as well the storage footprint, resiliency to privacy attacks, and descriptor maintenance.
As shown in \cref{tab:compare_sota_cambridge}, 
\camr{HLoc with SuperPoint and SuperGlue is the most accurate method but also has the highest storage requirements while being vulnerable to privacy attacks. Using HLoc with a newly developed descriptor method will require the map to be updated.
In end-to-end methods,}
DSAC++ is the most accurate method while being resilient to privacy attacks as it does not need to transmit visual descriptors. However, as it requires 4 model versions trained per-scene, it requires 828 MB storage to work under 4 scenes compared to our 48.12 MB. 
\camr{HybridSC as the most storage-efficient method keeps only 1.5\% if its original points via compression.}
However, it is unclear whether the privacy issue still remains for this method since it still relies on full visual descriptors to perform matching.
\camr{Notice, compressing scene structure can be theoretically combined with GoMatch to lower our storage requirements, which we leave as future work to design suitable scene compression techniques for geometric-base matching.
On the whole, GoMatch and MS-Transformer both properly balance those three aspects showing benefits in storage, privacy and absence of descriptor maintenance, and are competitive in accuracy.}
Compared to its visual-descriptor SuperPoint counterpart, GoMatch requires only 1.5\% of the capacity to store same scene. 
GoMatch reduces the average pose errors by $(10.67m, 95.7^{\circ})$ compared to our only prior geometric-based matching work, 
significantly reducing the accuracy gap to state-of-the-art methods.
We hope this inspires researchers to pursue this line of work.

\subsection{Generalization}
\label{sec:exp-generalization}

\begin{table}[t]
\caption{\newtext{Generalization study on 7-Scenes. GoMatch generalizes between different scene types and detector types and outperforming BPnPNet and PoseNet. } }
\label{tab:generalization}
\centering
\setlength{\tabcolsep}{4pt}
\resizebox{\textwidth}{!} {
\begin{tabular}{l l c c c c c c c c c c}
\toprule
&\multirow{2}{*}{Method} & Storage & No Desc. &Priv & Chess & Fire & Heads & Office & Pumpkin & Kitchen & Stairs \\ 
& &  (MB)  & Maint. & -acy & \multicolumn{7}{c}{Median Pose Error (m, $^{\circ}$) ($\downarrow$)}\\
\midrule
\multirow{3}*{\begin{sideways} E2E \end{sideways}} 
& PoseNet~\cite{Kendall2015ICCV}    & 350 & \camr{\cmark} & \cmark & 0.32/8.12 & 0.47/14.4 & 0.29/12.0 & 0.48/7.68 & 0.47/8.42 & 0.59/8.64 & 0.47/13.8 \\
& DSAC++~\cite{Brachmann2018CVPR}   & 1449 & \camr{\cmark} & \cmark & \textbf{0.02/0.50} & \textbf{0.02/0.90} & \textbf{0.01}/0.80 & \textbf{0.03/0.70} & \textbf{0.04/1.10} & \textbf{0.04/1.10} & 0.09/2.60 \\
& \camr{MSPN~\cite{Blanton2020CVPRW}}   & \camr{-} & \camr{\cmark} & \camr{\cmark} & \camr{0.09/4.76} & \camr{0.29/10.5 } & \camr{0.16/13.1} & \camr{0.16/6.8} & \camr{0.19/5.5} & \camr{0.21/6.61} & \camr{0.31/11.63} \\
& \camr{MS-Transformer~\cite{Shavit2021ICCV}}   & \camr{71.1} & \camr{\cmark} & \camr{\cmark} & \camr{0.11/4.66} & \camr{0.24/9.6} & \camr{0.14/12.19} & \camr{0.17/5.66} & \camr{0.18/4.44} & \camr{0.17/5.94} & \camr{0.26/8.45} \\
\midrule
\multirow{3}*{\begin{sideways} VM \end{sideways}} 
& Active Search~\cite{Sattler2017PAMI} & -     & \xmark & \xmark & 0.04/1.96 & 0.03/1.53 & 0.02/1.45 & 0.09/3.61 & 0.08/3.10 & 0.07/3.37 & \textbf{0.03}/2.22 \\
& \camr{HLoc~\cite{Sarlin2019CVPR}(w.SIFT~\cite{Lowe2004IJCV})} &        2923 &\xmark  &\xmark  & 0.03/1.13 & 0.03/1.08 & 0.02/2.19 & 0.05/1.42 & 0.07/1.80 & 0.06/1.84 & 0.18/4.41 \\
& \camr{HLoc(w.SP~\cite{Detone2018CVPRW})} & 22977 & \xmark  &\xmark  & 0.03/1.28 & 0.03/1.3 & 0.02/1.99 & 0.04/1.31 & 0.06/1.63 & 0.06/1.73 & 0.07/1.91\\
& \camr{HLoc(w.SP+SG~\cite{Sarlin2020CVPR})} & 22977 & \xmark  &\xmark  & \textbf{0.02}/0.85 & \textbf{0.02}/0.94 & \textbf{0.01/0.75} & \textbf{0.03}/0.92 & 0.05/1.30 & \textbf{0.04}/1.40 & 0.05/\textbf{1.47} \\
\midrule
\multirow{4}*{\begin{sideways} GM \end{sideways}} 
& BPnPNet~\cite{Campbell2020ECCV}(SIFT~\cite{Lowe2004IJCV}) & 302 & \cmark & \cmark & 1.29/43.82 & 1.48/51.82 & 0.93/55.13 & 2.61/59.06 & 2.15/39.85 & 2.15/43.00 & 2.98/60.27\\
& BPnPNet (SP~\cite{Detone2018CVPRW})   & 397 & \cmark & \cmark & 1.25/43.9 & 1.42/45.09 & 0.8/50.05 & 2.33/14.54 & 1.71/31.81 & 1.68/33.91 & 2.1/55.78\\
& GoMatch (SIFT)                        & 302 & \cmark & \cmark & 0.04/1.65 & 0.13/3.86 & 0.09/5.17 & 0.11/2.48 & 0.16/3.32 & 0.13/2.84 & 0.89/21.12\\
& GoMatch (SP)       & 397 & \cmark & \cmark & 0.04/1.56	& 0.12/3.71	& 0.05/3.43	& 0.07/1.76	& 0.28/5.65	& 0.14/3.03	& 0.58/13.12\\
\bottomrule
\end{tabular}
}
\end{table}

\newtext{As our final experiment, we study the generalization capability of our method in terms of localization in different types of scenes, \eg, indoor and outdoor, and matching keypoints obtained using different detectors. 
According to our results in \cref{tab:generalization}, similar to our previous expriments, we outperform BPnPNet by a large margin achieving $(1.43m, 34.7^{\circ})$ lower average median pose errors.
Except for GoMatch with SIFT keypoints which produces a relatively large $21.12^{\circ}$ median rotation error in Stairs, we are only slightly worse than our visual-based matching baselines with SIFT and SuperPoint.
Yet, we require only 10/1.7\% of the storage that is required by SIFT/SuperPoint to store maps.
We also largely outperform PoseNet~\cite{Kendall2015ICCV} in all metrics for all scenes except for the relatively lower translation error in Stairs scene, \ie, (0.47m vs 0.58m).
\camr{Furthermore, we achieve better pose than MS-Transformer in the majority of scenes, at the expense of a higher storage requirement.}
The results clearly verify that GoMatch trained on outdoor scenes (MegaDepth) generalizes smoothly to indoor scenes (7-Scenes), being agnostic to scene types.
Similarly, we also confirm that GoMatch trained with SIFT keypoints generalizes well to SuperPoint keypoints, being agnostic to detector types. 
}

\section{Conclusion}

We present GoMatch, a novel sparse keypoint matching method for visual localization that relies only on geometrical information and that carefully balances 
common practical challenges of large-scale localization, namely: localization performance, storage demands, privacy and
\camr{descriptor maintenance (or lack thereof)}. From all these, the last three are often overlooked.
Through a rigorous architecture design process, GoMatch dramatically surpasses its prior work in handling outliers, enabling it for real-world localization.
Compared to localization pipelines using visual descriptor-based matching, GoMatch allows localization with a minimal 3D scene representation, requiring as little as $1.5/1.7\%$ to store the same scene.
Geometric-based matching brings localization pipelines to a new level of scalability that opens the door for \newtext{localizing} in much larger environments.
We see our work as a starting point for this new direction and we look forward to inspire other researchers to pursue more accurate and reliable geometric-based visual localization in the future.

\PAR{Acknowledgments.}This research was partially funded by the Humboldt Foundation through the Sofja Kovalevskaya Award.

\clearpage
\bibliographystyle{ieee_fullname.bst}
\bibliography{main}

\begin{thebibliography}{10}\itemsep=-1pt

\bibitem{Arandjelovic2016CVPR}
Relja Arandjelovic, Petr Gronat, Akihiko Torii, Tomas Pajdla, and Josef Sivic.
\newblock Netvlad: Cnn architecture for weakly supervised place recognition.
\newblock In {\em IEEE Conference on Computer Vision and Pattern Recognition
  (CVPR)}, 2016.

\bibitem{Balntas2018ECCV}
Vassileios Balntas, Shuda Li, and Victor Prisacariu.
\newblock Relocnet: Continuous metric learning relocalisation using neural
  nets.
\newblock In {\em European Conference on Computer Vision (ECCV)}, September
  2018.

\bibitem{Bhowmik2020CVPR}
Aritra Bhowmik, Stefan Gumhold, Carsten Rother, and Eric Brachmann.
\newblock Reinforced feature points: Optimizing feature detection and
  description for a high-level task.
\newblock In {\em IEEE Conference on Computer Vision and Pattern Recognition
  (CVPR)}, pages 4948--4957, 2020.

\bibitem{Blanton2020CVPRW}
Hunter Blanton, Connor Greenwell, Scott Workman, and Nathan Jacobs.
\newblock Extending absolute pose regression to multiple scenes.
\newblock In {\em Proceedings of the IEEE/CVF Conference on Computer Vision and
  Pattern Recognition (CVPR) Workshops}, June 2020.

\bibitem{Brachmann2017CVPR}
Eric Brachmann, Alexander Krull, Sebastian Nowozin, Jamie Shotton, Frank
  Michel, Stefan Gumhold, and Carsten Rother.
\newblock {DSAC - Differentiable RANSAC for Camera Localization}.
\newblock In {\em IEEE Conference on Computer Vision and Pattern Recognition
  (CVPR)}, 2017.

\bibitem{Brachmann2018CVPR}
Eric Brachmann and Carsten Rother.
\newblock Learning less is more - 6d camera localization via 3d surface
  regression.
\newblock In {\em IEEE Conference on Computer Vision and Pattern Recognition
  (CVPR)}, 2018.

\bibitem{Brachmann2019CVPR}
Eric Brachmann and Carsten Rother.
\newblock Expert sample consensus applied to camera re-localization.
\newblock In {\em IEEE Conference on Computer Vision and Pattern Recognition
  (CVPR)}, pages 7525--7534, 2019.

\bibitem{Brachmann2019ICCV}
Eric Brachmann and Carsten Rother.
\newblock Neural-guided ransac: Learning where to sample model hypotheses.
\newblock In {\em IEEE International Conference on Computer Vision (ICCV)},
  pages 4322--4331, 2019.

\bibitem{OpenCV}
G. Bradski.
\newblock {The OpenCV Library}.
\newblock {\em Dr. Dobb's Journal of Software Tools}, 2000.

\bibitem{Brown2015ICCV}
Mark Brown, David Windridge, and Jean-Yves Guillemaut.
\newblock Globally optimal 2d-3d registration from points or lines without
  correspondences.
\newblock In {\em Proceedings of the IEEE International Conference on Computer
  Vision (ICCV)}, December 2015.

\bibitem{Campbell2020ECCV}
Dylan Campbell, Liu Liu, and Stephen Gould.
\newblock Solving the blind perspective-n-point problem end-to-end with robust
  differentiable geometric optimization.
\newblock In {\em European Conference on Computer Vision (ECCV)}, pages
  244--261. Springer, 2020.

\bibitem{Campbell2017ICCV}
Dylan Campbell, Lars Petersson, Laurent Kneip, and Hongdong Li.
\newblock Globally-optimal inlier set maximisation for simultaneous camera pose
  and feature correspondence.
\newblock In {\em Proceedings of the IEEE International Conference on Computer
  Vision (ICCV)}, Oct 2017.

\bibitem{Campbell2019CVPR}
Dylan Campbell, Lars Petersson, Laurent Kneip, Hongdong Li, and Stephen Gould.
\newblock The alignment of the spheres: Globally-optimal spherical mixture
  alignment for camera pose estimation.
\newblock In {\em Proceedings of the IEEE/CVF Conference on Computer Vision and
  Pattern Recognition (CVPR)}, June 2019.

\bibitem{Camposeco2019CVPR}
Federico Camposeco, Andrea Cohen, Marc Pollefeys, and Torsten Sattler.
\newblock Hybrid scene compression for visual localization.
\newblock In {\em Proceedings of the IEEE/CVF Conference on Computer Vision and
  Pattern Recognition (CVPR)}, June 2019.

\bibitem{Cao2014CVPR}
Song Cao and Noah Snavely.
\newblock Minimal scene descriptions from structure from motion models.
\newblock In {\em Proceedings of the IEEE Conference on Computer Vision and
  Pattern Recognition (CVPR)}, June 2014.

\bibitem{Cavallari20193DV}
Tommaso Cavallari, Luca Bertinetto, Jishnu Mukhoti, Philip Torr, and Stuart
  Golodetz.
\newblock Let's take this online: Adapting scene coordinate regression network
  predictions for online rgb-d camera relocalisation.
\newblock In {\em 2019 International Conference on 3D Vision (3DV)}, pages
  564--573, 2019.

\bibitem{Chelani2021CVPR}
Kunal Chelani, Fredrik Kahl, and Torsten Sattler.
\newblock How privacy-preserving are line clouds? recovering scene details from
  3d lines.
\newblock In {\em IEEE Conference on Computer Vision and Pattern Recognition
  (CVPR)}, pages 15668--15678, 2021.

\bibitem{Cheng2019ICCV}
Wentao Cheng, Weisi Lin, Kan Chen, and Xinfeng Zhang.
\newblock Cascaded parallel filtering for memory-efficient image-based
  localization.
\newblock In {\em Proceedings of the IEEE/CVF International Conference on
  Computer Vision (ICCV)}, October 2019.

\bibitem{Cuturi2013NIPS}
Marco Cuturi.
\newblock Sinkhorn distances: Lightspeed computation of optimal transport.
\newblock In C.J. Burges, L. Bottou, M. Welling, Z. Ghahramani, and K.Q.
  Weinberger, editors, {\em Advances in Neural Information Processing Systems},
  volume~26. Curran Associates, Inc., 2013.

\bibitem{David2004IJCV}
Philip David, Daniel Dementhon, Ramani Duraiswami, and Hanan Samet.
\newblock Softposit: Simultaneous pose and correspondence determination.
\newblock {\em International Journal of Computer Vision}, 59(3):259--284, 2004.

\bibitem{Deng2018CVPR}
Haowen Deng, Tolga Birdal, and Slobodan Ilic.
\newblock Ppfnet: Global context aware local features for robust 3d point
  matching.
\newblock In {\em Proceedings of the IEEE Conference on Computer Vision and
  Pattern Recognition (CVPR)}, June 2018.

\bibitem{Detone2018CVPRW}
Daniel DeTone, Tomasz Malisiewicz, and Andrew Rabinovich.
\newblock Superpoint: Self-supervised interest point detection and description.
\newblock In {\em CVPR Workshops}, pages 224--236, 2018.

\bibitem{Ding2019ICCV}
Mingyu Ding, Zhe Wang, Jiankai Sun, Jianping Shi, and Ping Luo.
\newblock Camnet: Coarse-to-fine retrieval for camera re-localization.
\newblock In {\em IEEE International Conference on Computer Vision (ICCV)},
  pages 2871--2880, 2019.

\bibitem{Dosovitskiy2016NeurIPS}
Alexey Dosovitskiy and Thomas Brox.
\newblock Generating images with perceptual similarity metrics based on deep
  networks.
\newblock In D. Lee, M. Sugiyama, U. Luxburg, I. Guyon, and R. Garnett,
  editors, {\em Advances in Neural Information Processing Systems}, volume~29.
  Curran Associates, Inc., 2016.

\bibitem{Dosovitskiy2016CVPR}
Alexey Dosovitskiy and Thomas Brox.
\newblock Inverting visual representations with convolutional networks.
\newblock In {\em IEEE Conference on Computer Vision and Pattern Recognition
  (CVPR)}, pages 4829--4837, 2016.

\bibitem{Dusmanu2021ICCV}
Mihai Dusmanu, Ondrej Miksik, Johannes~L Schonberger, and Marc Pollefeys.
\newblock Cross-descriptor visual localization and mapping.
\newblock In {\em IEEE International Conference on Computer Vision (ICCV)},
  pages 6058--6067, 2021.

\bibitem{Dusmanu2019CVPR}
Mihai Dusmanu, Ignacio Rocco, Tomas Pajdla, Marc Pollefeys, Josef Sivic,
  Akihiko Torii, and Torsten Sattler.
\newblock {D2-Net: A Trainable CNN for Joint Detection and Description of Local
  Features}.
\newblock In {\em IEEE Conference on Computer Vision and Pattern Recognition
  (CVPR)}, 2019.

\bibitem{Dusmanu2020CVPR}
Mihai Dusmanu, Johannes~L Sch{\"o}nberger, Sudipta~N Sinha, and Marc Pollefeys.
\newblock Privacy-preserving image features via adversarial affine subspace
  embeddings.
\newblock {\em IEEE Conference on Computer Vision and Pattern Recognition
  (CVPR)}, 2020.

\bibitem{Fischler81CACM}
Martin~A Fischler and Robert~C Bolles.
\newblock Random sample consensus: a paradigm for model fitting with
  applications to image analysis and automated cartography.
\newblock {\em CACM}, 24(6):381--395, 1981.

\bibitem{Gao2003TPAMI}
Xiao-Shan Gao, Xiao-Rong Hou, Jianliang Tang, and Hang-Fei Cheng.
\newblock Complete solution classification for the perspective-three-point
  problem.
\newblock {\em IEEE Transactions on Pattern Analysis and Machine Intelligence},
  25(8):930--943, 2003.

\bibitem{Geppert2020ECCV}
Marcel Geppert, Viktor Larsson, Pablo Speciale, Johannes~L Sch{\"o}nberger, and
  Marc Pollefeys.
\newblock Privacy preserving structure-from-motion.
\newblock In {\em European Conference on Computer Vision (ECCV)}, pages
  333--350. Springer, 2020.

\bibitem{Geppert2021CVPR}
Marcel Geppert, Viktor Larsson, Pablo Speciale, Johannes~L Schonberger, and
  Marc Pollefeys.
\newblock Privacy preserving localization and mapping from uncalibrated
  cameras.
\newblock In {\em IEEE Conference on Computer Vision and Pattern Recognition
  (CVPR)}, pages 1809--1819, 2021.

\bibitem{Germain2020ECCV}
Hugo Germain, Guillaume Bourmaud, and Vincent Lepetit.
\newblock S2dnet: Learning accurate correspondences for sparse-to-dense feature
  matching.
\newblock {\em European Conference on Computer Vision (ECCV)}, 2020.

\bibitem{Huang2021CVPR}
Shengyu Huang, Zan Gojcic, Mikhail Usvyatsov, Andreas Wieser, and Konrad
  Schindler.
\newblock Predator: Registration of 3d point clouds with low overlap.
\newblock In {\em Proceedings of the IEEE/CVF Conference on Computer Vision and
  Pattern Recognition (CVPR)}, pages 4267--4276, June 2021.

\bibitem{Ke2017CVPR}
Tong Ke and Stergios~I. Roumeliotis.
\newblock An efficient algebraic solution to the perspective-three-point
  problem.
\newblock In {\em Proceedings of the IEEE Conference on Computer Vision and
  Pattern Recognition (CVPR)}, July 2017.

\bibitem{Kendall2016ICRA}
Alex Kendall and Roberto Cipolla.
\newblock Modelling uncertainty in deep learning for camera relocalization.
\newblock In {\em IEEE International Conference on Robotics and Automation
  (ICRA)}, 2016.

\bibitem{Kendall2017CVPR}
Alex Kendall and Roberto Cipolla.
\newblock {Geometric Loss Functions for Camera Pose Regression With Deep
  Learning}.
\newblock In {\em IEEE Conference on Computer Vision and Pattern Recognition
  (CVPR)}, 2017.

\bibitem{Kendall2015ICCV}
Alex Kendall, Matthew Grimes, and Roberto Cipolla.
\newblock Posenet: A convolutional network for real-time 6-dof camera
  relocalization.
\newblock In {\em IEEE International Conference on Computer Vision (ICCV)},
  2015.

\bibitem{Kingma2015ICLR}
Diederik~P. Kingma and Jimmy Ba.
\newblock Adam: A method for stochastic optimization.
\newblock In {\em ICLR (Poster)}, 2015.

\bibitem{Kneip2011CVPR}
Laurent Kneip, Davide Scaramuzza, and Roland Siegwart.
\newblock A novel parametrization of the perspective-three-point problem for a
  direct computation of absolute camera position and orientation.
\newblock In {\em IEEE Conference on Computer Vision and Pattern Recognition
  (CVPR)}, 2011.

\bibitem{Laskar2017ICCVW}
Zakaria Laskar, Iaroslav Melekhov, Surya Kalia, and Juho Kannala.
\newblock {Camera Relocalization by Computing Pairwise Relative Poses Using
  Convolutional Neural Network}.
\newblock In {\em IEEE International Conference on Computer Vision (ICCV)
  Workshops}, Oct 2017.

\bibitem{Li2020CVPR}
Xiaotian Li, Shuzhe Wang, Yi Zhao, Jakob Verbeek, and Juho Kannala.
\newblock Hierarchical scene coordinate classification and regression for
  visual localization.
\newblock In {\em IEEE Conference on Computer Vision and Pattern Recognition
  (CVPR)}, pages 11983--11992, 2020.

\bibitem{Li2018CVPR}
Zhengqi Li and Noah Snavely.
\newblock Megadepth: Learning single-view depth prediction from internet
  photos.
\newblock In {\em IEEE Conference on Computer Vision and Pattern Recognition
  (CVPR)}, 2018.

\bibitem{Liu2020arXiv}
Liu Liu, Dylan Campbell, Hongdong Li, Dingfu Zhou, Xibin Song, and Ruigang
  Yang.
\newblock Learning 2d-3d correspondences to solve the blind perspective-n-point
  problem, 2020.

\bibitem{Lowe2004IJCV}
David~G. Lowe.
\newblock Distinctive image features from scale-invariant keypoints.
\newblock {\em International Journal of Computer Vision}, 60(2):91--110, Nov
  2004.

\bibitem{Luo2020CVPR}
Zixin Luo, Lei Zhou, Xuyang Bai, Hongkai Chen, Jiahui Zhang, Yao Yao, Shiwei
  Li, Tian Fang, and Long Quan.
\newblock Aslfeat: Learning local features of accurate shape and localization.
\newblock In {\em IEEE Conference on Computer Vision and Pattern Recognition
  (CVPR)}, pages 6589--6598, 2020.

\bibitem{MeraTrujillo20203DV}
Marcela Mera-Trujillo, Benjamin Smith, and Victor Fragoso.
\newblock Efficient scene compression for visual-based localization.
\newblock In {\em 2020 International Conference on 3D Vision (3DV)}, pages
  1--10, 2020.

\bibitem{Moo2018CVPR}
Kwang Moo~Yi, Eduard Trulls, Yuki Ono, Vincent Lepetit, Mathieu Salzmann, and
  Pascal Fua.
\newblock Learning to find good correspondences.
\newblock In {\em IEEE Conference on Computer Vision and Pattern Recognition
  (CVPR)}, pages 2666--2674, 2018.

\bibitem{MorenoNoguerECCV2008}
Francesc Moreno-Noguer, Vincent Lepetit, and Pascal Fua.
\newblock Pose priors for simultaneously solving alignment and correspondence.
\newblock In {\em European Conference on Computer Vision (ECCV)}, pages
  405--418. Springer, 2008.

\bibitem{Muja2014PAMI}
Marius Muja and David~G. Lowe.
\newblock Scalable nearest neighbor algorithms for high dimensional data.
\newblock {\em IEEE Transactions on Pattern Analysis and Machine Intelligence},
  36(11):2227--2240, 2014.

\bibitem{Ng2022CVPR}
Tony Ng, Hyo~Jin Kim, Vincent~T Lee, Daniel DeTone, Tsun-Yi Yang, Tianwei Shen,
  Eddy Ilg, Vassileios Balntas, Krystian Mikolajczyk, and Chris Sweeney.
\newblock Ninjadesc: Content-concealing visual descriptors via adversarial
  learning.
\newblock In {\em IEEE Conference on Computer Vision and Pattern Recognition
  (CVPR)}, pages 12797--12807, 2022.

\bibitem{Pittaluga2019CVPR}
Francesco Pittaluga, Sanjeev~J Koppal, Sing~Bing Kang, and Sudipta~N Sinha.
\newblock Revealing scenes by inverting structure from motion reconstructions.
\newblock In {\em IEEE Conference on Computer Vision and Pattern Recognition
  (CVPR)}, pages 145--154, 2019.

\bibitem{Radenovic2018TPAM}
Filip Radenovi{\'c}, Giorgos Tolias, and Ond{\v{r}}ej Chum.
\newblock Fine-tuning cnn image retrieval with no human annotation.
\newblock {\em TPAM}, 41(7):1655--1668, 2018.

\bibitem{Radwan2018RAL}
Noha Radwan, Abhinav Valada, and Wolfram Burgard.
\newblock Vlocnet++: Deep multitask learning for semantic visual localization
  and odometry.
\newblock {\em IEEE Robotics and Automation Letters}, 3(4):4407--4414, 2018.

\bibitem{Sarlin2019CVPR}
Paul-Edouard Sarlin, Cesar Cadena, Roland Siegwart, and Marcin Dymczyk.
\newblock From coarse to fine: Robust hierarchical localization at large scale.
\newblock In {\em IEEE Conference on Computer Vision and Pattern Recognition
  (CVPR)}, 2019.

\bibitem{Sarlin2020CVPR}
Paul-Edouard Sarlin, Daniel DeTone, Tomasz Malisiewicz, and Andrew Rabinovich.
\newblock Superglue: Learning feature matching with graph neural networks.
\newblock In {\em IEEE Conference on Computer Vision and Pattern Recognition
  (CVPR)}, pages 4938--4947, 2020.

\bibitem{Sarlin2021CVPR}
Paul-Edouard Sarlin, Ajaykumar Unagar, Mans Larsson, Hugo Germain, Carl Toft,
  Viktor Larsson, Marc Pollefeys, Vincent Lepetit, Lars Hammarstrand, Fredrik
  Kahl, et~al.
\newblock Back to the feature: Learning robust camera localization from pixels
  to pose.
\newblock In {\em IEEE Conference on Computer Vision and Pattern Recognition
  (CVPR)}, pages 3247--3257, 2021.

\bibitem{Sattler2017PAMI}
Torsten Sattler, Bastian Leibe, and Leif Kobbelt.
\newblock Efficient \& effective prioritized matching for large-scale
  image-based localization.
\newblock {\em IEEE Transactions on Pattern Analysis and Machine Intelligence},
  39(9):1744--1756, 2017.

\bibitem{Sattler2018CVPR}
Torsten Sattler, Will Maddern, Carl Toft, Akihiko Torii, Lars Hammarstrand,
  Erik Stenborg, Daniel Safari, Masatoshi Okutomi, Marc Pollefeys, Josef Sivic,
  et~al.
\newblock Benchmarking 6dof outdoor visual localization in changing conditions.
\newblock In {\em IEEE Conference on Computer Vision and Pattern Recognition
  (CVPR)}, pages 8601--8610, 2018.

\bibitem{Sattler2017CVPR}
Torsten Sattler, Akihiko Torii, Josef Sivic, Marc Pollefeys, Hajime Taira,
  Masatoshi Okutomi, and Tomas Pajdla.
\newblock {Are Large-Scale 3D Models Really Necessary for Accurate Visual
  Localization?}
\newblock In {\em IEEE Conference on Computer Vision and Pattern Recognition
  (CVPR)}, 2017.

\bibitem{Sattler2019CVPR}
Torsten Sattler, Qunjie Zhou, Marc Pollefeys, and Laura Leal-Taixe.
\newblock Understanding the limitations of cnn-based absolute camera pose
  regression.
\newblock In {\em IEEE Conference on Computer Vision and Pattern Recognition
  (CVPR)}, 2019.

\bibitem{Schoenberger2016CVPR}
Johannes~Lutz Sch\"{o}nberger and Jan-Michael Frahm.
\newblock Structure-from-motion revisited.
\newblock In {\em IEEE Conference on Computer Vision and Pattern Recognition
  (CVPR)}, 2016.

\bibitem{Schoenberger2018CVPR}
Johannes~L Sch{\"o}nberger, Marc Pollefeys, Andreas Geiger, and Torsten
  Sattler.
\newblock {Semantic Visual Localization}.
\newblock In {\em IEEE Conference on Computer Vision and Pattern Recognition
  (CVPR)}, 2018.

\bibitem{Schonberger2016ECCV}
Johannes~L Sch{\"o}nberger, Enliang Zheng, Jan-Michael Frahm, and Marc
  Pollefeys.
\newblock Pixelwise view selection for unstructured multi-view stereo.
\newblock In {\em European Conference on Computer Vision (ECCV)}, pages
  501--518. Springer, 2016.

\bibitem{Shavit2021ICCV}
Yoli Shavit, Ron Ferens, and Yosi Keller.
\newblock Learning multi-scene absolute pose regression with transformers.
\newblock In {\em Proceedings of the IEEE/CVF International Conference on
  Computer Vision (ICCV)}, pages 2733--2742, October 2021.

\bibitem{Shotton2013CVPR}
Jamie Shotton, Ben Glocker, Christopher Zach, Shahram Izadi, Antonio Criminisi,
  and Andrew Fitzgibbon.
\newblock Scene coordinate regression forests for camera relocalization in
  rgb-d images.
\newblock In {\em IEEE Conference on Computer Vision and Pattern Recognition
  (CVPR)}, pages 2930--2937, 2013.

\bibitem{Sinkhorn1967PJM}
Richard Sinkhorn and Paul Knopp.
\newblock Concerning nonnegative matrices and doubly stochastic matrices.
\newblock {\em Pacific Journal of Mathematics}, 21(2):343--348, 1967.

\bibitem{Speciale2019CVPR}
Pablo Speciale, Johannes~L Schonberger, Sing~Bing Kang, Sudipta~N Sinha, and
  Marc Pollefeys.
\newblock Privacy preserving image-based localization.
\newblock In {\em IEEE Conference on Computer Vision and Pattern Recognition
  (CVPR)}, pages 5493--5503, 2019.

\bibitem{Sun2021CVPR}
Jiaming Sun, Zehong Shen, Yuang Wang, Hujun Bao, and Xiaowei Zhou.
\newblock Loftr: Detector-free local feature matching with transformers.
\newblock In {\em IEEE Conference on Computer Vision and Pattern Recognition
  (CVPR)}, pages 8922--8931, 2021.

\bibitem{Sun2020CVPR}
Weiwei Sun, Wei Jiang, Eduard Trulls, Andrea Tagliasacchi, and Kwang~Moo Yi.
\newblock Acne: Attentive context normalization for robust
  permutation-equivariant learning.
\newblock In {\em Proceedings of the IEEE/CVF Conference on Computer Vision and
  Pattern Recognition (CVPR)}, June 2020.

\bibitem{Taira2018CVPR}
Hajime Taira, Masatoshi Okutomi, Torsten Sattler, Mircea Cimpoi, Marc
  Pollefeys, Josef Sivic, Tomas Pajdla, and Akihiko Torii.
\newblock {InLoc: Indoor Visual Localization with Dense Matching and View
  Synthesis}.
\newblock In {\em IEEE Conference on Computer Vision and Pattern Recognition
  (CVPR)}, 2018.

\bibitem{Thomas2019ICCV}
Hugues Thomas, Charles~R. Qi, Jean-Emmanuel Deschaud, Beatriz Marcotegui,
  Francois Goulette, and Leonidas~J. Guibas.
\newblock Kpconv: Flexible and deformable convolution for point clouds.
\newblock In {\em Proceedings of the IEEE/CVF International Conference on
  Computer Vision (ICCV)}, October 2019.

\bibitem{Toft2020TPAMI}
Carl Toft, Will Maddern, Akihiko Torii, Lars Hammarstrand, Erik Stenborg,
  Daniel Safari, Masatoshi Okutomi, Marc Pollefeys, Josef Sivic, Tomas Pajdla,
  Fredrik Kahl, and Torsten Sattler.
\newblock Long-term visual localization revisited.
\newblock {\em IEEE Transactions on Pattern Analysis and Machine Intelligence},
  44(4):2074--2088, 2022.

\bibitem{Torii2015CVPR}
Akihiko Torii, Relja Arandjelovic, Josef Sivic, Masatoshi Okutomi, and Tomas
  Pajdla.
\newblock 24/7 place recognition by view synthesis.
\newblock In {\em IEEE Conference on Computer Vision and Pattern Recognition
  (CVPR)}, pages 1808--1817, 2015.

\bibitem{TranTIP2019}
Ngoc-Trung Tran, Dang-Khoa Le~Tan, Anh-Dzung Doan, Thanh-Toan Do, Tuan-Anh Bui,
  Mengxuan Tan, and Ngai-Man Cheung.
\newblock On-device scalable image-based localization via prioritized cascade
  search and fast one-many ransac.
\newblock {\em IEEE Transactions on Image Processing}, 28(4):1675--1690, 2019.

\bibitem{Ulyanov2017arXiv}
Dmitry Ulyanov, Andrea Vedaldi, and Victor Lempitsky.
\newblock Instance normalization: The missing ingredient for fast stylization,
  2017.

\bibitem{VaswaniNeurIPS2017}
Ashish Vaswani, Noam Shazeer, Niki Parmar, Jakob Uszkoreit, Llion Jones,
  Aidan~N Gomez, \L~ukasz Kaiser, and Illia Polosukhin.
\newblock Attention is all you need.
\newblock In I. Guyon, U.~V. Luxburg, S. Bengio, H. Wallach, R. Fergus, S.
  Vishwanathan, and R. Garnett, editors, {\em Advances in Neural Information
  Processing Systems}, volume~30. Curran Associates, Inc., 2017.

\bibitem{Walch2017ICCV}
Florian Walch, Caner Hazirbas, Laura Leal-Taixe, Torsten Sattler, Sebastian
  Hilsenbeck, and Daniel Cremers.
\newblock Image-based localization using lstms for structured feature
  correlation.
\newblock In {\em IEEE International Conference on Computer Vision (ICCV)},
  2017.

\bibitem{Wang2020ECCV}
Qianqian Wang, Xiaowei Zhou, Bharath Hariharan, and Noah Snavely.
\newblock Learning feature descriptors using camera pose supervision.
\newblock In {\em European Conference on Computer Vision (ECCV)}, 2020.

\bibitem{Wang2019ACMTG}
Yue Wang, Yongbin Sun, Ziwei Liu, Sanjay~E. Sarma, Michael~M. Bronstein, and
  Justin~M. Solomon.
\newblock Dynamic graph cnn for learning on point clouds.
\newblock {\em ACM Trans. Graph.}, 38(5), Oct. 2019.

\bibitem{Yang2019CVPR}
Luwei Yang, Ziqian Bai, Chengzhou Tang, Honghua Li, Yasutaka Furukawa, and Ping
  Tan.
\newblock Sanet: Scene agnostic network for camera localization.
\newblock In {\em IEEE Conference on Computer Vision and Pattern Recognition
  (CVPR)}, pages 42--51, 2019.

\bibitem{Zhang2019ICCV}
Jiahui Zhang, Dawei Sun, Zixin Luo, Anbang Yao, Lei Zhou, Tianwei Shen, Yurong
  Chen, Long Quan, and Hongen Liao.
\newblock Learning two-view correspondences and geometry using order-aware
  network.
\newblock In {\em Proceedings of the IEEE/CVF International Conference on
  Computer Vision (ICCV)}, October 2019.

\bibitem{Zhou2020ICRA}
Qunjie Zhou, Torsten Sattler, Marc Pollefeys, and Laura Leal-Taixe.
\newblock To learn or not to learn: Visual localization from essential
  matrices.
\newblock In {\em IEEE International Conference on Robotics and Automation
  (ICRA)}, pages 3319--3326. IEEE, 2020.

\end{thebibliography}

\clearpage
\appendix
In this supplementary material, we provide additional insights and details about certain aspects of the main paper that were not fundamental for its understanding, but that are helpful for interested readers, to fully comprehend and replicate our method. In \cref{sec:arch_details}, we provide in-depth descriptions of each network component with mathematical definitions as well as architecture diagrams.
Next, we present dataset preparation details including dataset processing, query   sampling and correspondence ground truth generation in \cref{sec:dataset_details}, followed by the implementation details regarding our training procedure, hyper-parameter choices and an overview of inference process in \cref{sec:impl_details}.
\newtext{In \cref{sec:metrics}, we provide additional details on how to compute the metrics used in the paper, in particular the mean reprojection error Area Under the cumulative Curve (AUC).}
\camr{Finally, in \cref{sec:exp-challenges} we detail how we generate \cref{tab:storage_plans} and provide more insights about the practical challenges of large-scale structure-based localization.}

\section{Architecture Details} \label{sec:arch_details}

\begin{figure}
    \centering
    \includegraphics[width=0.45\textwidth]{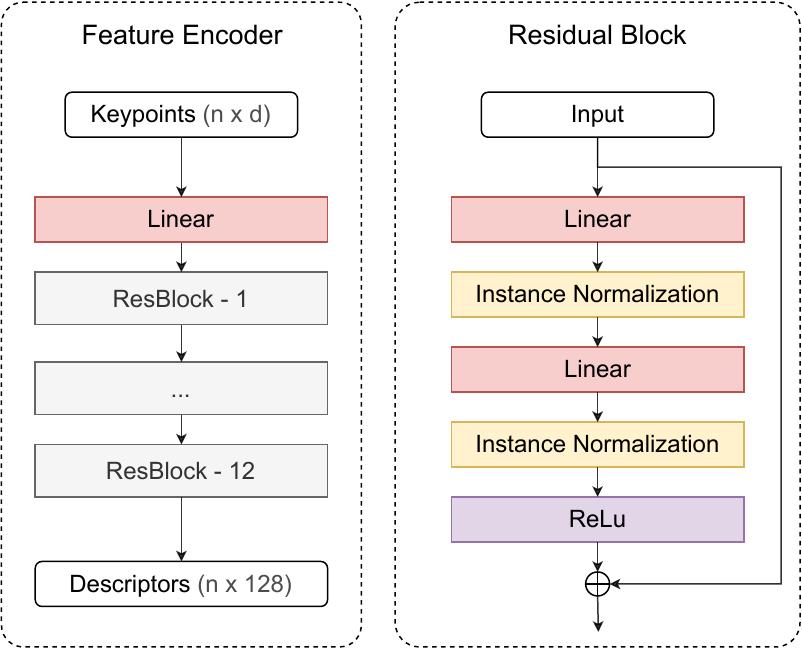}
    \caption{Feature Encoder (left) \& Residual Block (right)}
    \label{fig:encoder}
\end{figure}

\PAR{Feature Encoder.} 
We adopt the same encoder architecture as the one presented in ~\cite{Campbell2020ECCV,Moo2018CVPR}, which is shown in \cref{fig:encoder}.
The encoder (\textit{left}) consists of a cascade of residual blocks, where each block (\textit{right}) is composed of sequential point-shared fully connected layers, followed by instance normalization and non-linearity (ReLUs).
\newtext{We use two siamese encoders with shared weight parameters to encode bearing vectors of both 2D image keypoints and 3D points, since it has been shown to improve GoMatch's performance in \cref{sec:exp-ablations}.}

\PAR{Attention.}
We use a graph neural network to update features from the same modality (\textit{self-attention}) and multi-head attention layers to exchange features across modalities (\textit{cross-attention}).
This is similar to PREDATOR~\cite{Huang2021CVPR}, although in that work, the attention module is inserted in the bottleneck of the keypoint encoder, \ie, KPConv~\cite{Thomas2019ICCV}, and thus it is applied to the downsampled keypoints.
In our case, the feature encoder does not have a bottleneck and we instead apply the attention directly to the final output of our encoder.
The full attention module contains self $\to$ cross $\to$ self-attention layers and is applied to both query and database encoded features.

We employ an independent graph neural network~\cite{Wang2019ACMTG} to the features coming from each modality. These exchange and refine features of every keypoint type with a fixed number of closest neighbors in coordinate space. Let $\vec f_i \in \R^d$ represent the feature for keypoint $i$ and $\vec f_j$ the feature corresponding to one of its neighbors. Let $\mathcal{E}_i$ represent the set of edges formed between keypoint $i$ to its closest $K$ neighbors in coordinate space. We update feature $f_i$ according to
\begin{align}
    \left.\vec f^{(k+1)}_i\right|_{k = \{0, 1\}} &= \operatorname*{max}_{j:(i,j) \in \mathcal{E}_i} h_\theta \left(\operatorname{cat}[\vec f^{(k)}_i, \vec f^{(k)}_j - \vec f^{(k)}_i] \right)\\
    \vec f^{(3)}_i &= h_\theta \left(\operatorname{cat}[\vec f^{(0)}_i, \vec f^{(1)}_i, \vec f^{(2)}_i] \right),
\end{align}
where $h_\theta$ is a composition of a fully connected layer, instance normalization~\cite{Ulyanov2017arXiv}, and a leaky ReLU, and $\vec f^{(0)}_i$ is the feature produced by the encoder for keypoint $i$.

To exchange information across modalities we leverage cross-attention~\cite{VaswaniNeurIPS2017}. In a cross-attention layer, every keypoint in a modality will interact with keypoints from the other modality through multi-head attention. There are three important concepts that govern this interaction: queries, keys and values. This setup mimics traditional maps or dictionaries, albeit in a continuous form, where every key is paired with a value and these values can be extracted by querying them with the appropriate key. Specifically, given a feature $\vec f_i$ coming from a keypoint in one modality and features $\vec g_j$ from an arbitrary keypoint $j$ in the other modality, we form the query, keys and values according to $\vec q_i = \mat W_q \vec f_i$, $\vec k_j = \mat W_k \vec g_j$ and $\vec v_j = \mat W_v \vec g_j$, where $\mat W_q, \mat W_k, \mat W_v \in \R^{d\times d}$ are parameters learned by the network. We update  feature $\vec f_i$ according to
\begin{align}
    \alpha_{ij} &= \operatorname{softmax}(\vec q_i^\top \vec k_j / \sqrt{d}) \\
    \vec m_i &= \sum_j \alpha_{ij} \vec v_j \\
    \vec f^{(k+1)}_i &= \vec f^{(k)}_i + \operatorname{MLP}(\operatorname{cat}[\vec q_i, \vec m_i]).
\end{align}

\PAR{Sinkhorn Matching.}
After getting refined features from the attention module, the Sinkhorn matching stage is responsible for assigning correspondences between query and database keypoints. To that effect, we leverage the Sinkhorn algorithm~\cite{Cuturi2013NIPS,Sinkhorn1967PJM} that has foundations in optimal transport theory, to solve the assignment problem with holistic reasoning. Sinkhorn produces a joint discrete probability distribution of two keypoints being matched.
Consider the following features after attention $\vec f_i^{\text{ATT}}$ and $\vec g_j^{\text{ATT}}$, corresponding to the i-th query keypoint and the j-th database keypoint. We construct an assignment cost matrix $\mat M \in \R^{M\times N}$ as
\begin{align}
     m_{ij} = \left\|\frac{\vec f_i^{\text{ATT}}}{\|\vec f_i^{\text{ATT}}\|} - \frac{\vec g_j^{\text{ATT}}}{\|\vec g_j^{\text{ATT}}\|}\right\|,
\end{align}
where $m_{ij}$ is the element in the i-th row and j-th column of $\mat M$. %
Sinkhorn provides a solution for the following entropy regularized optimization problem:
\begin{equation}
    \mat P^* = \argmin_{\mat P \in \mathcal{U}(\vec r, \vec c)} \sum_{i=1}^M \sum_{j=1}^N m_{ij} p_{ij} - \tau p_{ij}\log p_{ij}\\
\end{equation}
where $\mathcal{U}(\vec r, \vec c) := \{\mat P \in \R_+^{M\times N}, \mat P \1_N = \vec c, \mat P^\top \1_M = \vec r\}$, $\vec r \in R^M_+$, $\vec c \in R^N_+$, $\sum_i^M r_i = 1$, $\sum_j^N c_j = 1$ and with $\1_M$ denoting a vector of 1s of size $M$.
The vectors $\vec r$ and $\vec c$ represent the (marginal) probability vectors of keypoints being matched. In our setup, these marginals are initialized uniformly. The hyperparameter $\tau > 0$ controls the strength of the regularization.
This problem is solved iteratively and in a differentiable way through successive steps of row and column-wise normalization, as presented in Cuturi~\cite{Cuturi2013NIPS}.

In a realistic scenario, many of these keypoints will not have a match and in the worst case scenario, it can happen that all points from both sets are unmatched. To handle that, we update the matching cost matrix $\mat M$ with an extra row and column that act as a ``gutter" for unmatched points and denote this new matrix $\tilde{\mat M} \in \R^{M+1 \times N+1}$. We also add an extra element to both row and column marginals, that is able to ``absorb" all all unmatched points in the other set if needed, forming the augmented marginals $\tilde{\vec r} \in \R^{M+1}_+$ and $\tilde{\vec c} \in \R^{N+1}_+$. We present the updated cost matrix and marginals below
\begin{align}
    \tilde{\vec r} = \begin{bmatrix} \frac{1}{M + N} \\ \vdots \\ \frac{1}{M + N} \\ \frac{N}{M + N}\end{bmatrix} & \begin{bmatrix} m_{11} & \dots & m_{1N} & m_u\\
    \vdots & \ddots & \vdots & \vdots \\
    m_{M1} & \dots & m_{MN} & m_u\\
    m_u & \dots & m_u & m_u
    \end{bmatrix} = \tilde{\mat M}\\
    & \begin{bmatrix} \frac{1}{M + N} & \dots & \frac{1}{M + N} & \frac{M}{M + N}\end{bmatrix} = \tilde{\vec c},
\end{align}
where $m_u$ is a parameter learned by the network that represents the cost of considering points as unmatched. With the cost and marginals as input, Sinkhorn generates the final discrete probability distribution $\tilde{\mat P} \in \R^{M+1 \times N+1}_+$, representing soft correspondences. The pose estimation pipeline requires hard correspondences so we retain only pairs of keypoints that mutually assign to each other.

\begin{figure*}
    \centering
    \includegraphics[width=0.95\textwidth]{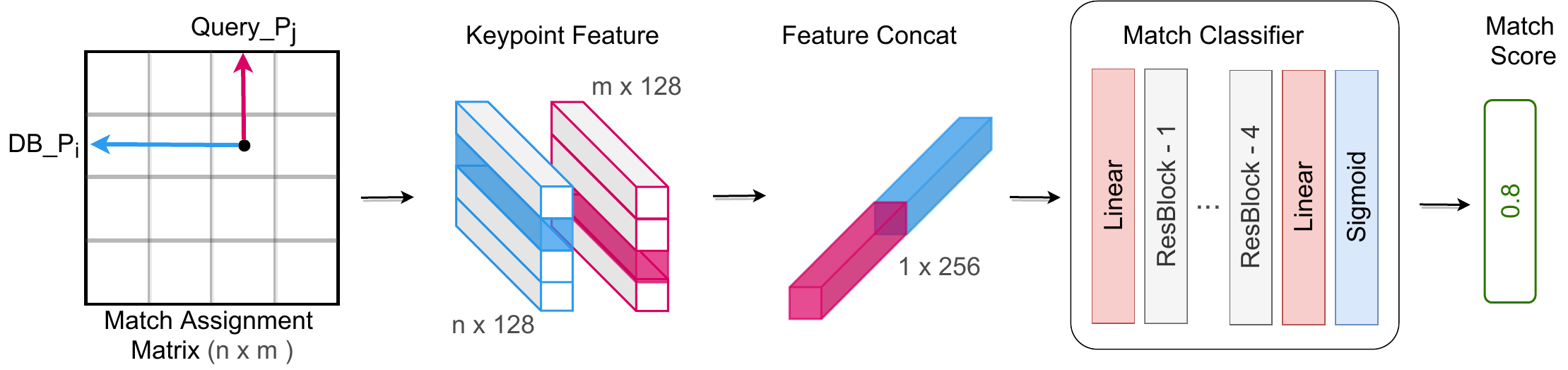}
    \caption{Outlier rejection through classification. For a match identified by Sinkhorn, we collect the matched keypoint features that are extracted by the encoder and attention module. The feature for a query keypoint and a database keypoint is concatenated into a single feature representing the match. The match classifier then predicts the match confidence (as a probability score) which can be used later to filter a less confident match.}
    \label{fig:outlier-rejection}
\end{figure*}

\PAR{Outlier Rejection.} 
After Sinkhorn matching, the estimated corresponding pairs may still contain outlier matches. 
We follow \cite{Moo2018CVPR} to cast the outlier rejection as a binary classification task, where we use a classifier to predict a confidence score to identify whether a match is an inlier or outlier \wrt a specific confidence threshold.
As depicted in \cref{fig:outlier-rejection}, given a predicted match referring to a query keypoint $Query\_P_j$ and a database keypoint $DB\_P_i$, we first collect keypoint feature (extracted in the previous stage by the encoder and attention module).
We then concatenate the features of its involved keypoints to form a single feature representation of that match and feed it into a classification network.
The classifier follows the overall architecture of keypoint encoder described in \cref{fig:encoder} and has a final classification layer, \ie, a linear layer and a sigmoid operation, to output a probability score for an input match which is then used to filter a less confident match.

\section{Dataset Details}\label{sec:dataset_details}
\PAR{MegaDepth~\cite{Li2018CVPR}.}
Our networks are trained on MegaDepth. We first followed the preprocessing steps from \cite{Dusmanu2019CVPR} to \newtext{generate the undistorted reconstructions}.
We adopted the test set published by MegaDepth's authors, composed of 53 sequences.
We further split the remaining sequences into training and validation splits, where 
the training and validation splits ensure a similar distribution in terms of sequence sizes between both. 
\newtext{Within each data split, we first sample up to 500 queries per scene. For each query, we collect its $k$ co-visible views that have at least 35\% of visual overlapping, where we drop queries with not enough co-visible views, \ie, $k<3$.
The visual overlapping is computed by the number of commonly seen 3D points divided by the total seen 3D points in the query view. 
Notice, in practice, those co-visible views are supposed to be obtained by image retrieval techniques~\cite{Arandjelovic2016CVPR,Radenovic2018TPAM,Torii2015CVPR} (as we do for evaluating on the other two benchmark datasets). 
However, for training and ablation study, we use ground-truth information provided by the dataset to guarantee the co-visibility, which enables stable training and allows us to focus on analysing the geometric-based matching performance with proper retrieval quality.}

\newtext{In total, our training set contains 18881 queries covering 99 scenes, our validation set contains 3146 queries covering 16 scenes and our testing set contains samples 7344 covering 49 scenes.}

\PAR{Cambridge Landmarks~\cite{Kendall2015ICCV}.}
\newtext{
Cambridge Landmarks is an outdoor urban localization dataset, where each frame is annotated with a camera pose label.
The video footage was captured through a smartphone camera and includes significant urban clutter from pedestrian and vehicles. 
The dataset is composed of five scenes and each scene is composed of multiple sequences. We report results in four of these scenes -- King's College, Old Hospital, Shop Facade and St. Mary's Church -- amounting to a total of 29 sequences, that in our case were all used for testing. }
\newtext{
To obtain 3D points for our method, we use the publicly available reconstruction generated per-scene with SuperPoint~\cite{Detone2018CVPRW} from the training images and ground truth poses.
This reconstruction was originally made available by Torsten Sattler and has been used in the recent work PixLoc~\cite{Sarlin2021CVPR} for localization evaluation.
In addition, we also follow PixLoc to use their released top-10 query-retrieval pairs computed by NetVLAD~\cite{Arandjelovic2016CVPR}.
Both reconstruction and retrieval pairs are hosted \href{https://cvg-data.inf.ethz.ch/pixloc_CVPR2021/Cambridge-Landmarks}{here} by the authors of PixLoc.
}

\PAR{7-Scenes~\cite{Shotton2013CVPR}.}
7-Scenes is a pose annotated indoor dataset of seven different scenes, captured with an RGB-D camera. Each scene is composed of multiple sequences and every frame in each sequence comes with a color image, a depth image, and the camera pose.
\newtext{The dataset has a total of 46 sequences and we used 18 for testing, following the original test split.
We use the top-10 query-retrieval pairs computed by DenseVLAD~\cite{Torii2015CVPR} which were made available \href{https://cvg-data.inf.ethz.ch/pixloc_CVPR2021/7Scenes/7scenes_densevlad_retrieval/}{here} by the authors of PixLoc~\cite{Sarlin2021CVPR} and used in their experiments.
}
\newtext{To obtain reference 3D points from database retrievals, we }
first use a keypoint detector on the color image to generate an original set of candidate 2D keypoints. 
We then transform their coordinates from the color image space to depth image space and only retain keypoints that have a valid depth measurement, rounding fractional coordinates to the nearest pixel location. 
Given a 2D image location and the corresponding depth value, we compute a 3D keypoint in camera space and then transform it to scene's frame of reference. 
For simplicity and enabled by the fact that we only make use of a sparse subset of points in the original depth image, we don't try to establish 3D point correspondences between different co-visible frames and consider that each depth image observes a set of unique 3D points in each frame. 
\newtext{This process allows us to flexibly generate 3D points using different types of keypoint. As shown in Section 5.6, we generated two versions of 3D points that are obtained using SIFT~\cite{Lowe2004IJCV} and SuperPoint~\cite{Deng2018CVPR} keypoints to demonstrate the generalization capability of GoMatch.}

\PAR{Keypoint Detection.}
To be consistent with the 3D models in MegaDepth that are reconstructed with SIFT~\cite{Lowe2004IJCV} keypoints, we also use a SIFT detector to extract keypoints from query images. 
We limit all detectors to extract up to 1024 keypoints per image, this is to fit our model into a 12GB GPU during testing. 
On 7-Scenes, we are able to apply any keypoint detectors, since we have the depth map to extract the corresponding 3D coordinate.
In our experiments, we used a hand-crafted detector -- SIFT -- and a state-of-the-art learning-based detector -- SuperPoint~\cite{Detone2018CVPRW} -- to extract keypoints.
\newtext{For Cambridge Landmarks, we also used SuperPoint, ensuring consistency with the reconstructed 3D model.}
All of the keypoints are pre-computed for both datasets and cached locally.

\PAR{Ground Truth Correspondences.} \label{par:gt-correspondences}
To train our network with the matching loss \newtext{and to compute reprojection-based AUC metric (\cf \cref{sec:metrics})}, we need to generate ground truth (gt) correspondence labels between query keypoints and 3D point cloud keypoints (database keypoints). 
Given a set of database keypoints and a query image with its known camera pose, we project the 3D points into the query image to obtain its 2D projections.
Then we perform mutual nearest neighbour search based on the L2 distance between \newtext{the query keypoints and the projected 3D keypoints and consider them as a gt correspondence if the distance is below a threshold of \camr{0.001} in normalized image coordinates, i.e., distance between bearing vectors. 
}

\section{Implementation Details}
\label{sec:impl_details}
\PAR{Architecture.}  We used 12 residual blocks for keypoint encoders  (\cf \cref{fig:encoder}) and 4 residual blocks for match classifier for outlier rejection (\cf \cref{fig:outlier-rejection}). The features produced by the keypoint encoders and attention module have dimension 128. The graph neural network used for \textit{self attention} establishes a KNN graph with the closest 10 neighbors, in coordinate space. We use 4 parallel heads for multi-head \textit{cross attention}. 
\newtext{In total, GoMatch has approximately 1.3M weight parameters.}

\PAR{Training.}  We train all models using the ADAM~\cite{Kingma2015ICLR} optimizer at the learning rate of $0.001$. The batch sizes are chosen to allow each model to be trained on a single 48GB NVIDIA Quadro RTX 8000 GPU and vary from \newtext{16 to 64 batches (depending on the model size), \eg, we use batch size 16 for GoMatch. } 
We train each model for $50$ epochs and determine the best checkpoint based on the lowest loss value on the validation set. 
\newtext{GoMatch requires approximately 20 hours training time for 50 epochs.}

\PAR{Outlier Rate Control.} \newtext{The keypoints involved in gt matches are the inlier keypoints and the remaining keypoints are considered outliers.} For training stability, we constrain the keypoint outlier rate to $0.5$. Besides our ablation studies \newtext{shown in \cref{fig:orate_abalt}}, no keypoint outlier control is applied during testing. 

\PAR{Keypoint Number Control.} 
By default, we limit query and database keypoints to a maximum number of 1024, to ensure the testing can be performed on a 12GB GPU.
For query keypoints, we enforce the detector to extract at most 1024 keypoints.
For database keypoints, we use up to its first 1024 keypoints if exceeding that number.
During training, we ignore training samples where the number of query keypoints or database keypoints is less than 100.
\newtext{During inference, we consider a sample as failed if the number of query keypoints or database keypoints is less than 10.}

\PAR{Inference.} Given a query image and a 3D point cloud, we first extract 2D keypoints from the query using detectors such as SIFT~\cite{Lowe2004IJCV} or SuperPoint~\cite{Detone2018CVPRW} (in practice, we load pre-computed keypoints).
Next, we identify $k$ \newtext{retrieval/}co-visible reference views to compute 3D points that are visible to query. Correspondences are established between the query keypoints and co-visible 3D points (in bearing vectors) by GoMatch. 
We run the Sinkhorn algorithm for a maximum of 20 iterations to obtain an initial  set of match estimates.
Then we remove all matches with classification scores below a threshold 0.5 to filter uncertain matches and finally estimate camera pose of the query. 

\PAR{Pose Estimation.} We use the OpenCV~\cite{OpenCV} library to estimate pose from a set of query to point cloud keypoint correspondences. We first identify an initial set of inlier matches using a minimal P3P~\cite{Gao2003TPAMI} solver paired with RANSAC~\cite{Fischler81CACM} and then estimate the final pose through a Levenberg-Marquardt optimization step acting only on the inliers. We allow RANSAC to perform 1000 iterations with  the admissible maximum inlier error threshold for the bearing vectors set to 1e-3.

\section{Reprojection Error AUC}\label{sec:metrics}
\begin{figure}[t]
    \centering
    \includegraphics[width=0.6\textwidth]{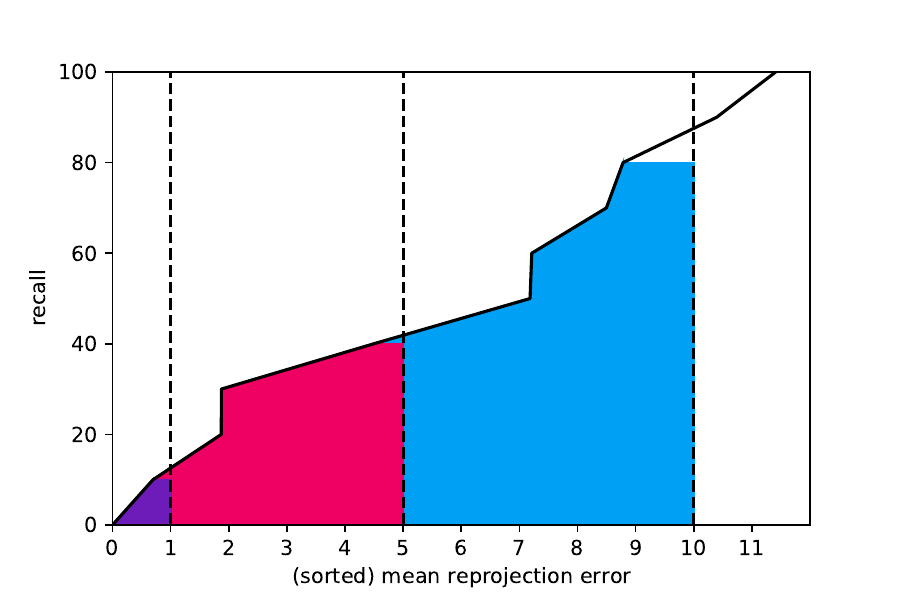}
    \caption{A hypothetical example of how the mean reprojection error AUC is computed for the thresholds at 1/5/10 pixels. 
    After sorting the mean reprojection error and establishing a bijective relation with the recall, one computes the area under the curve using the trapezoidal rule of integration.
    Despite not visible, the AUC in cyan for the 10 pixel threshold spans from 0 to 10, and similar to the pink area for the 5 pixels threshold.
    }
    \label{fig:reprojection-error-auc}
\end{figure}

\newtext{
For MegaDepth, we reported the mean reprojection error area under the cummulative curve (AUC) with thresholds at 1, 5 and 10 pixels.
This metric was inspired by the pose error area under the cummulative curve used in \cite{Sarlin2020CVPR,Sarlin2021CVPR,Sun2021CVPR}.
However, in our case, as the scale unit of MegaDepth is undetermined and might be inconsistent across scenes, we have to compute the AUC based on reprojection error instead of pose errors to ensure consistency. 
To compute our proposed AUC metric, for each query, we project the inlier 3D points, \ie, the 3D points involved in the ground-truth (gt) correspondences, onto the query image using the gt pose and the estimated pose.
The mean reprojection errors are computed as the mean pixel distances between the gt projections and estimated projections of the 3D points.
Then we compute the area under the cummulative recall curve up to a specific pixel error threshold as illustrated in \cref{fig:reprojection-error-auc}.
Finally, we normalize the area by the error threshold to keep the metric score  $\in [0, 100]$.

As described in \cref{sec:dataset_details}, the gt correspondences are computed with \camr{a 0.001 normalized} threshold, which means the
\camr{bearing vectors} obtained by projecting the inlier 3D points using the gt camera pose have up to \camr{a 0.001} distance to \camr{the bearing vectors of} our labelled gt 2D keypoints.
\camr{This tolerance is reflected in pixel space, resulting in}
relatively low \camr{AUC} scores at 1 pixel threshold (as shown in \cref{tab:ablat_archs}) even using our Oracle matcher.
However, this does not affect the function of the AUC metric that is to reveal the performance gap between different methods.
We showed that the AUC metric yields similar conclusion as the pose error quantile metric used by BPnPNet~\cite{Campbell2020ECCV}.
}

\section{Practical Challenges in Large-scale Localization}\label{sec:exp-challenges}

As introduced in \camr{the main paper}, the primary motivation of our work is to present a new localization framework that does not suffer from: (i) storage requirements, (ii) descriptor maintenance effort~\cite{Dusmanu2021ICCV}, and (iii) privacy concerns~\cite{Pittaluga2019CVPR, Dusmanu2020CVPR}. These are challenges that current structure-based localization methods face, especially when it comes to scaling-up for city-level scenes.
A considerable amount of literature focuses on making localization methods more accurate, however, the aforementioned practical challenges are much less explored.
To study the storage requirements, we perform a detailed analysis on MegaDepth, a dataset that resembles a city-level large-scale environment. While MegaDepth is a collection of landmarks instead of a real city-scale scene, we argue that a city-scale scene consists of a collection of districts (similar to landmarks).
As shown in \camr{\cref{tab:storage_plans}}, the minimal scene data, \ie, scene coordinates (3D) and camera metadata (Cameras), is always required by a structured-based localization to obtain 2D-3D correspondences, which will take 15.73MB and 3.44GB storage in total for all Megadepth scenes. This is the only data that geometric-based matching (GM) methods need to store.
This is in contrast to visual-based matching (VM), as they need to store extra visual descriptors that consume 130/1040GB for SIFT/SuperPoint descriptors.
Alternatively, one can extract descriptors on-the-fly from the retrieved raw images, which requires saving all of the raw images with an extra storage of 157GB and leading to more computational burden.
\textbf{Then}, after addressing the storage issue, one still needs to consider the privacy vulnerability raised during descriptor transmission, since large-scale localization with a 3D map has to be realistically deployed in a server-client mode, where the server stores 3D scene data required to perform localization. This is another motivation driving us towards using geometric information only.
\textbf{Finally}, as thoroughly covered in~\cite{Dusmanu2021ICCV}, with the continuous advance of local features, upgrading descriptors is inevitable in the long-term, and it involves either re-building the map or transforming the descriptors. In contrast, for geometric-based matching, upgrading our localization algorithm does not need an update on the map side unless the scene has changed, a case in which every map would need to be updated.

\end{document}